\documentclass[10pt,twocolumn,letterpaper]{article}

\usepackage[pagenumbers]{cvpr} 

\usepackage{fontawesome}

\usepackage{xspace}
\usepackage{multirow}
\setlist[itemize]{itemsep=0pt,topsep=0pt,partopsep=0pt,parsep=0pt,leftmargin=5mm}

\newcommand{\name}{ActionMesh\xspace}
\newcommand{\projectwebpage}{\textbf{\href{https://remysabathier.github.io/actionmesh/}{https://remysabathier.github.io/actionmesh/}}}

\DeclareMathOperator{\iattn}{infattn}
\DeclareMathOperator{\sattn}{selfattn}

\DeclareMathOperator{\reshape}{reshape}

\DeclareMathOperator*{\argmin}{argmin}

\newcommand{\encthreed}{\mathcal{E}_{\text{3D}}}
\newcommand{\decthreed}{\mathcal{D}_{\text{3D}}}
\newcommand{\decfourd}{\mathcal{D}_{\text{4D}}}
\newcommand{\genthreed}{\mathcal{G}_{\text{3D}}}
\newcommand{\mesh}{\mathcal{M}}
\newcommand{\meshref}{\mathcal{M}}
\newcommand{\verts}{\mathbf{V}}
\newcommand{\faces}{\mathbf{F}}
\newcommand{\deform}{\delta}
\newcommand{\img}{\mathbf{I}}
\newcommand{\latentfourd}{\mathbf{Z}}
\newcommand{\latentthreed}{\mathbf{z}}
\newcommand{\tokenfourd}{\mathbf{X}}
\newcommand{\pc}{\mathbf{P}}
\newcommand{\pcq}{\mathbf{Q}}
\newcommand{\pcs}{\mathbf{S}}

\newcommand{\nsrc}{N_{\text{S}}}
\newcommand{\ntgt}{N_\text{T}}
\newcommand{\npc}{N_\text{p}}
\newcommand{\cwindow}{c_\text{w}}
\def\RR{{\rm I\!R}}

\usepackage{microtype}

\definecolor{cvprblue}{rgb}{0.21,0.49,0.74}
\usepackage[pagebackref,breaklinks,colorlinks,allcolors=cvprblue]{hyperref}

\title{
\name: Animated 3D Mesh Generation with Temporal 3D Diffusion
}

\author{
Remy Sabathier$^{1,3}$ \hspace{2em} David Novotny$^2$ \hspace{2em} Niloy J. Mitra$^3$ \hspace{2em} Tom Monnier$^1$\\
$^1$Meta Reality Labs \hspace{2em} $^2$SpAItial \hspace{2em} $^3$University College London
}

\begin{document}

\twocolumn[{%
\renewcommand\twocolumn[1][]{#1}%
\maketitle
\vspace*{-.4in}
\begin{center}
    \projectwebpage
\end{center}
\begin{center}
\includegraphics[width=1.0\linewidth]{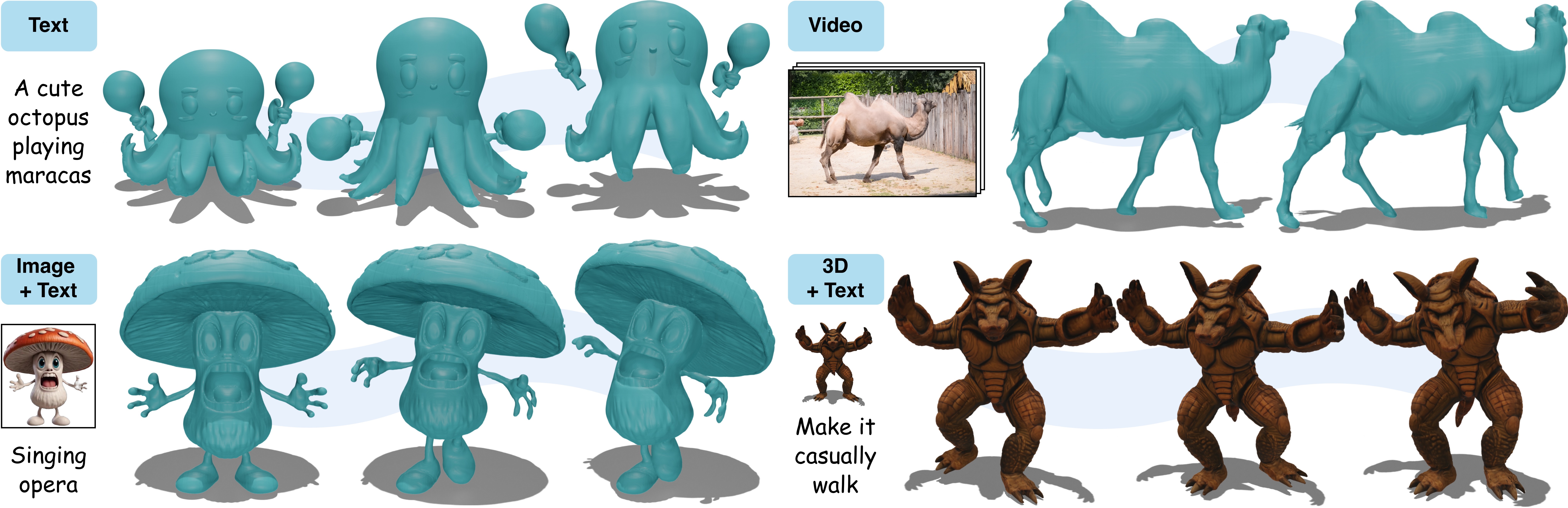} 
    \captionof{figure}{
\textbf{\name.} Our model generates 3D meshes `in action' from a wide range of inputs such as a text prompt, a video, an image + an animation text prompt, or a 3D mesh + an animation text prompt. Unlike previous approaches, our method is not only \textbf{fast}, but also \textbf{rig-free} and \textbf{topology consistent}. These properties are convenient in practice, \eg, they allow the seamless animation of complex 3D shapes like an octopus with maracas (top left) or the automatic transfer of the mesh texture throughout the animation (bottom right).
    }
    \label{fig:teaser}
\end{center}
\vspace{1.5em}
}]

\begin{abstract}

Generating animated 3D objects is at the heart of many applications, yet most advanced works are typically difficult to apply in practice because of their limited setup, their long runtime, or their limited quality. We introduce \name, a generative model that predicts production-ready 3D meshes `in action' in a feed-forward manner. Drawing inspiration from early video models, our key insight is to modify existing 3D diffusion models to include a temporal axis, resulting in a framework we dubbed `temporal 3D diffusion'. Specifically, we first adapt the 3D diffusion stage to generate a sequence of synchronized latents representing time-varying and independent 3D shapes. Second, we design a temporal 3D autoencoder that translates a sequence of independent shapes into the corresponding deformations of a pre-defined reference shape, allowing us to build an animation. Combining these two components, \name generates animated 3D meshes from different inputs like a monocular video, a text description, or even a 3D mesh with a text prompt describing its animation. Besides, compared to previous approaches, our method is fast and produces results that are rig-free and topology consistent, hence enabling rapid iteration and seamless applications like texturing and retargeting. We evaluate our model on the existing video-to-4D benchmark Consistent4D and introduce ActionBench, a new benchmark based on Objaverse, reporting state-of-the-art results in both geometric accuracy and temporal consistency.

\end{abstract}

\section{Introduction}
\label{sec:intro}

The ability to automatically produce animated 3D objects from simple user inputs is a core computer vision problem that holds high promise for any 3D-content application, like video games, animated movies and commercials, or augmented/virtual reality. 
However, despite recent progress, most works share three main limitations.
 First, they are specific to limited setups with a predefined input modality (\eg, a video) and predefined object categories (\eg, bipeds or rig-able objects). 
Second, they often rely on long (30-45 minutes) optimization loops, which are slow and prone to local minima.
Third, the overall output quality does not meet the production criteria.

In this paper, we introduce \emph{\name}, a feed-forward generative model that is simple, scalable, and computes production-ready 3D meshes `in action' from diverse inputs. At its core, 
\name is a new video-to-4D model that predicts an animated 3D mesh given a video as input.
Drawing inspiration from early video models, we extend existing 3D diffusion models with a temporal axis and separate the 3D generation from the animation prediction.
Our model runs in two stages.
First, we derive a \emph{temporal 3D diffusion model} from a pretrained 3D latent diffusion model, which produces a sequence of synchronized latents representing time-varying but independent 3D meshes.
Second, we design a \emph{temporal 3D autoencoder}, which converts a generic sequence of 3D meshes into the deformations of a chosen reference mesh, yielding an animation with constant topology. 
Importantly, for both stages, we build upon models with strong 3D priors to balance for the lack of 3D animated data.

Compared to existing approaches, \name is fast (2 minutes for 16-frame video)
and produces meshes that are rig-free and topology consistent. 
These properties are convenient in practice since (i)~it allows the animation of complex 3D shapes where rigging is unknown or difficult, and (ii)~it automatically preserves the mesh attributes, like a texture, throughout the animation.
In addition, thanks to its design, we show that our model can be extended to other generative tasks like text-to-4D, \{image+text\}-to-4D or \{3D+text\}-to-animation, as well as related applications like retargeting/motion transfer, as illustrated in \Cref{fig:teaser} and \Cref{fig:motion_transfer}. 
We evaluate our video-to-4D model qualitatively on the Consistent4D benchmark~\cite{jiang_consistent4d_2023} and quantitatively on ActionBench, a newly introduced Objaverse~\cite{deitke_objaverse_2022} benchmark.
Comparing against state-of-the-art methods, including DreamMesh4D~\cite{li2024dreammesh4d}, LIM~\cite{sabathier2025lim}, V2M4~\cite{chen2025v2m4}, and the concurrent work ShapeGen4D~\cite{yenphraphai2025shapegen4d}, we show that \name consistently outperforms competitors on both geometric accuracy and correspondence quality while achieving a speed-up of roughly $10\times$. Code and pretrained weights are available on our webpage: \textbf{\projectwebpage}.

\paragraph{Summary.} Our main contributions are three-fold:
\begin{itemize}
\item A fast feed-forward model called \name that generates animated 3D meshes from diverse inputs with an unprecedented speed and quality. 
\item A temporal 3D diffusion model that produces synchronized shape latents by minimally extending pretrained 3D diffusion backbones. 
\item A temporal 3D autoencoder that assembles independent shapes into a single topology-consistent animation via the deformation of a reference mesh.
\end{itemize}
\section{Related Work}
\label{sec:related_work}

\paragraph{3D foundational models.}

The seminal work of Zhang~\etal~\cite{10.1145/3592442}
introduces 3DShape2VecSet, a neural field supported by a set of latent vectors (a vecset), designed for scalable 3D encoding of meshes and point clouds, and explicitly targeting generative diffusion. 
Trellis~\citep{xiang2024structured} generalizes this idea to structured 3D latents, scaling to large models that can decode into multiple 3D representations such as radiance fields, Gaussian splats, and meshes. 
Building on vecset-style latents, Craftsman~\citep{li2024craftsman} performs 3D-native diffusion over latent sets coupled with a geometry refiner, enabling high-quality mesh synthesis and editing; Dora-VAE~\citep{Chen_2025_Dora} improves VAE sampling on similar representations with a strong emphasis on sharp edges and high-frequency details.
CLAY~\citep{zhang2024clay} further advances controllable 3D generation with a large 3D-native latent DiT for geometry and a dedicated diffusion model for materials.
In parallel, LRM~\citep{hong2024lrmlargereconstructionmodel} and follow-up works like~\citep{tang2024lgmlargemultiviewgaussian,siddiqui2024assetgen,zarzar24twinner} demonstrate that large transformers trained on massive image collections can reconstruct high-fidelity 3D NeRFs or Gaussians from a single view.
Among recent image-to-3D mesh generators, Hunyuan3D~\citep{lai2025hunyuan3d25highfidelity3d} predicts geometry with a flow-based DiT and texture with a separate model, generating high-resolution assets from text or images; TripoSG~\citep{li2025triposg} uses a large rectified-flow transformer for high-fidelity mesh reconstruction.
While image-to-3D models provide powerful frame-level priors, they process each instance independently and do not explicitly model temporal correspondences, making it difficult to produce an animated mesh with a constant topology across frames.

\begin{figure*}[t!]
\centering
\includegraphics[width=.98\linewidth]{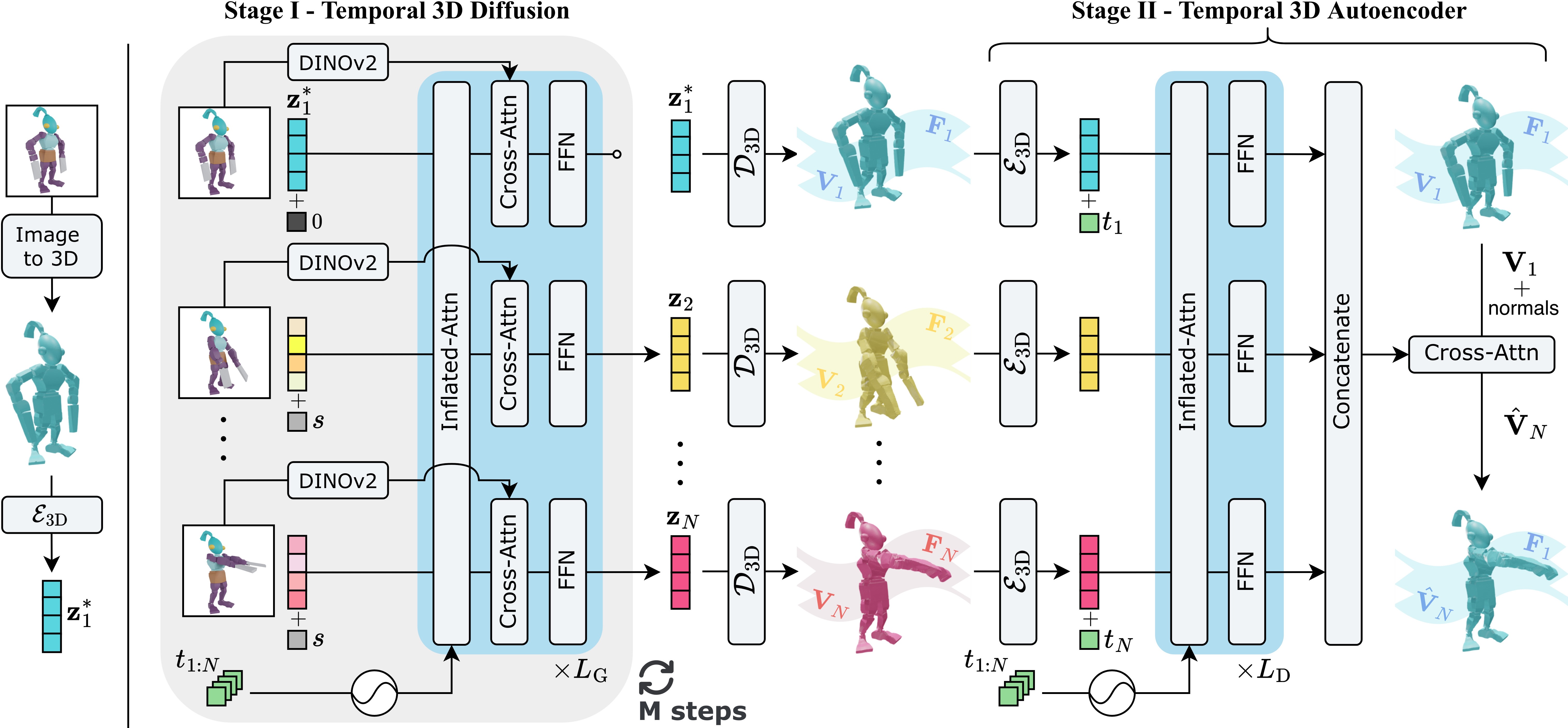}
\caption{\textbf{Overview.} Given an input video, our model generates an animated 3D mesh in two stages. \textbf{(Stage I)} After computing a reference mesh latent $\latentthreed_1^{*}$ with an off-the-shelf image-to-3D model, we use our temporal 3D diffusion model to produce from $\latentthreed_1^{*}$ and the video, a sequence of time-varying but independent 3D meshes. \textbf{(Stage II)} Our temporal 3D autoencoder takes theses shapes as input and predicts, for each shape in the sequence, a deformation field of the reference mesh vertices, thus yielding an animated 3D mesh with consistent topology.}
\label{fig:architecture}
\end{figure*}

\paragraph{Video-to-4D via optimization.}

Targeting dynamic 4D content, a common strategy first synthesizes multi-view or multi-frame videos from a monocular sequence and then optimizes a 4D representation.
SV4D and SV4D 2.0~\citep{Xie2024SV4DD3,Yao2025SV4D2E} unify multi-frame, multi-view video diffusion to supervise dynamic NeRFs, significantly improving spatio-temporal consistency but still requiring per-scene optimization.
CAT4D~\citep{Wu2024CAT4DCA} similarly converts a monocular video into multi-view sequences and then optimizes deformable 3D Gaussians, offering strong reconstructions and controllable camera and time, but without vertex-to-vertex correspondences on a single topology. 
Related approaches~\citep{jiang_consistent4d_2023, zeng_stag4d_2024,wu_sc4d_2024, yin_4dgen_2024, zhang2024fourdiffusion, ren_dreamgaussian4d_2023, wang2024vidu4dsinglegeneratedvideo} also rely on diffusion or generative priors to supervise dynamic Gaussians or NeRFs, typically with temporal regularizers or sparse controls and a subsequent optimization or training stage for each scene. 
DreamMesh4D~\citep{li2024dreammesh4d} predicts meshes via a mesh–Gaussian hybrid with sparsely controlled deformation followed by optimization. 
V2M4~\citep{chen2025v2m4} enforces topology and texture consistency through a multi-stage pipeline that includes camera search, reposing, pairwise registration, and global texture optimization.
LIM~\citep{sabathier2025lim} learns to interpolate a 3D implicit field over time and extracts a UV-textured mesh sequence with consistent topology through a test-time optimization. 
These methods achieve high-quality 4D reconstructions but rely on post-optimization or per-scene training.

\paragraph{Feed-forward 4D reconstruction.}

A complementary line of work reconstructs or generates 4D content in a single forward pass, avoiding test-time optimization. 
Motion2VecSets~\citep{Wei2024M2V} proposes a 4D diffusion model that denoises compressed latent sets for dynamic surface reconstruction from point-cloud sequences.
L4GM~\citep{ren2024l4gm} predicts a sequence of 3D Gaussian splats from a monocular video in one pass and then upsamples them temporally via a learned interpolation model, but still operates entirely in Gaussian space rather than producing an animated mesh with fixed topology.
Similarly, 4DGT~\citep{xu20254dgt} learns a transformer that directly predicts 4D Gaussians from real monocular videos, again operating in Gaussian space instead of explicit meshes. 
These existing feed-forward methods either produce Gaussians or neural fields instead of a mesh with constant topology. 
Concurrently, ShapeGen4D~\citep{yenphraphai2025shapegen4d} extends a pre-trained 3D generative model by introducing temporal attention, enabling feed-forward prediction of video-conditioned 4D meshes.
Unlike our approach, it does not explicitly enforce a single, globally consistent mesh topology across the sequence.

\paragraph{Animation-ready 4D assets.} 

Several works focus on making existing 3D assets \emph{animation-ready} by predicting rigs, skinning weights, or deformation fields, instead of reconstructing geometry from raw videos.
Make-It-Animatable~\citep{Guo_2025_CVPR} predicts skeletons, skinning weights, and pose rectification for meshes. 
MagicArticulate~\citep{Song_2025_CVPR} similarly generates articulation-ready rigs and deformations for a wide range of shapes. 
SMF~\cite{muralikrishnan2025smftemplatefreerigfreeanimation} proposes kinetic codes to semantically 
encode motion using a self-supervised framework.
RigAnything~\citep{Liu2025RigAnythingTA} extends auto-rigging to diverse object categories via an autoregressive transformer that predicts hierarchical skeletons and skinning weights. 
For Gaussian-based assets, RigGS~\citep{yao2025riggsrigging3dgaussians} recovers skeletons and skinning directly from videos, enabling articulated motion in the space of 3D Gaussians.
Several works~\citep{shi2025drive,wu2025animateanymeshfeedforward4dfoundation,jiang2024animate3danimating3dmodel} deform a mesh from a video or a text prompt, using either feed-forward diffusion models or test-time optimization.

These methods operate focus on rigging and animation, rather than reconstructing an animated mesh from a video. 
\section{Method}
\label{sec:method}

Our goal is to generate animated 3D meshes from various user inputs.
Instead of building input-specific models, we cast multiple generative tasks into the core \emph{video-to-4D} problem, whose goal is to generate an animated 3D mesh given a single video. 
To address this, we harness pretrained 3D generators to balance for the lack of 3D animated data.
Specifically, we build upon the 3D latent diffusion framework of 3DShape2VecSet~\citep{10.1145/3592442} (\Cref{sec:preliminaries}) and propose minimal modifications leading to our two-stage model called \emph{\name}.
First, we introduce a temporal 3D diffusion model, predicting a sequence of time-varying but independent 3D meshes (\Cref{sec:sync_denoiser}).
Second, we present a temporal 3D autoencoder, converting a sequence of independent 3D shapes into an animated mesh with a constant topology (\Cref{sec:fourd_reconstructor}).
Finally, we discuss how to apply our model to other scenarios in~\Cref{sec:applications}.
~\Cref{fig:architecture} shows an overview of \name.

\paragraph{Terminology and problem setting.}
We use `3D mesh' to describe a triangular 3D mesh denoted by \(\mesh = (\verts, \faces)\), where \(\verts \in \RR^{N_{\textrm{v}} \times 3}\) are the vertex positions and \(\faces \in \{1, ..., N_{\textrm{V}}\}^{N_{\textrm{f}} \times 3}\) captures the face connectivity. 
We use `4D mesh' to describe a sequence of time-varying but \emph{independent} 3D meshes that do not share the same topology, which we denote by \(\{(\verts_k, \faces_k)\}_{k=1}^{N}\).
Finally, we call `animated 3D mesh' a sequence of 3D meshes sharing the \emph{same} topology and denoted by \(\{(\verts_k, \faces)\}_{k=1}^{N}\).
Note that an animated 3D mesh is a particular form of a 4D mesh.

Let $\{\img_{k}\}_{k=1}^{N}$ be a video, where $\img_{k} \in \RR^{H \times W \times 3}$ is an RGB frame depicting an object in motion at framestep $t_k$.
Our objective is to predict an animated 3D mesh corresponding to the input video. Specifically, we aim at predicting a reference mesh $\meshref = (\verts, \faces)$ 
as well as its vertex position updates $\verts_k$ for each framestep $t_k$, such that $\verts_k$ represents the new vertices of $\meshref$ matching the motion depicted by $\img_{k}$.

\subsection{Background: 3DShape2VecSet}
\label{sec:preliminaries}

Following the paradigm of latent diffusion models~\cite{rombach22high-resolution}, \emph{3DShape2VecSet} learns a 3D diffusion model comprising two stages: 
(i) a variational autoencoder (VAE) made of an encoder $\encthreed$ and a decoder $\decthreed$ to encode 3D shapes into a compact latent space, and (ii) a diffusion model $\genthreed$ predicting latents conditioned on an input image.
Specifically, given a dense set of points $\pc$ sampled on the surface, a sparse set of query vectors $\pcq$ (called vector set or \emph{VecSet}, which is either learned or subsampled from $\pc$) is encoded by a cross-attention layer using $\pc$ as context.
Then, the query vectors are passed to several self-attention layers to produce the low-dimensional latent $\latentthreed$.
The decoder $\decthreed$ takes the latent $\latentthreed$, processes it with several self-attention layers and uses the output as context to a cross-attention layer producing the occupancy (or signed distance) of a query 3D point. 
Finally, a mesh is computed by querying the decoder on a dense grid of points and running a meshification algorithm like marching cubes~\cite{10.1145/37402.37422}.
Note that this architecture is reminiscent of Perceiver IO~\cite{Jaegle2021PerceiverGP} used in NLP.

The generative model $\genthreed$ is a diffusion transformer~\citep{Peebles2022DiT}, which takes the form of a decoder-only transformer~\citep{10.5555/3295222.3295349} where each block has an extra cross-attention layer to incorporate the conditioning signal.
In follow-up works, the conditioning signal is typically an image that is described using a frozen DINOv2~\citep{oquab2023dinov2}.
Many advanced image-to-3D models build on top of VecSet, \eg, CLAY~\citep{zhang2024clay}, Craftsman~\citep{li2024craftsman}, TripoSG~\citep{li2025triposg} or Hunyuan3D~\citep{lai2025hunyuan3d25highfidelity3d}. 
In this work, we adopt TripoSG as our backbone for its state-of-the-art performance and its open-source implementation, but our formulation should be applicable to other VecSet backbones. One specificity of TripoSG is that it relies on rectified flow~\citep{liu23flow,lipman22flow} and that the flow timestep $s \in [0,1000]$ is Fourier-embedded and concatenated as an additional token (see \cite{li2025triposg}).

\subsection{Stage I: Temporal 3D Diffusion}
\label{sec:sync_denoiser}

\begin{figure}[t!]
\centering
\includegraphics[width=\columnwidth]{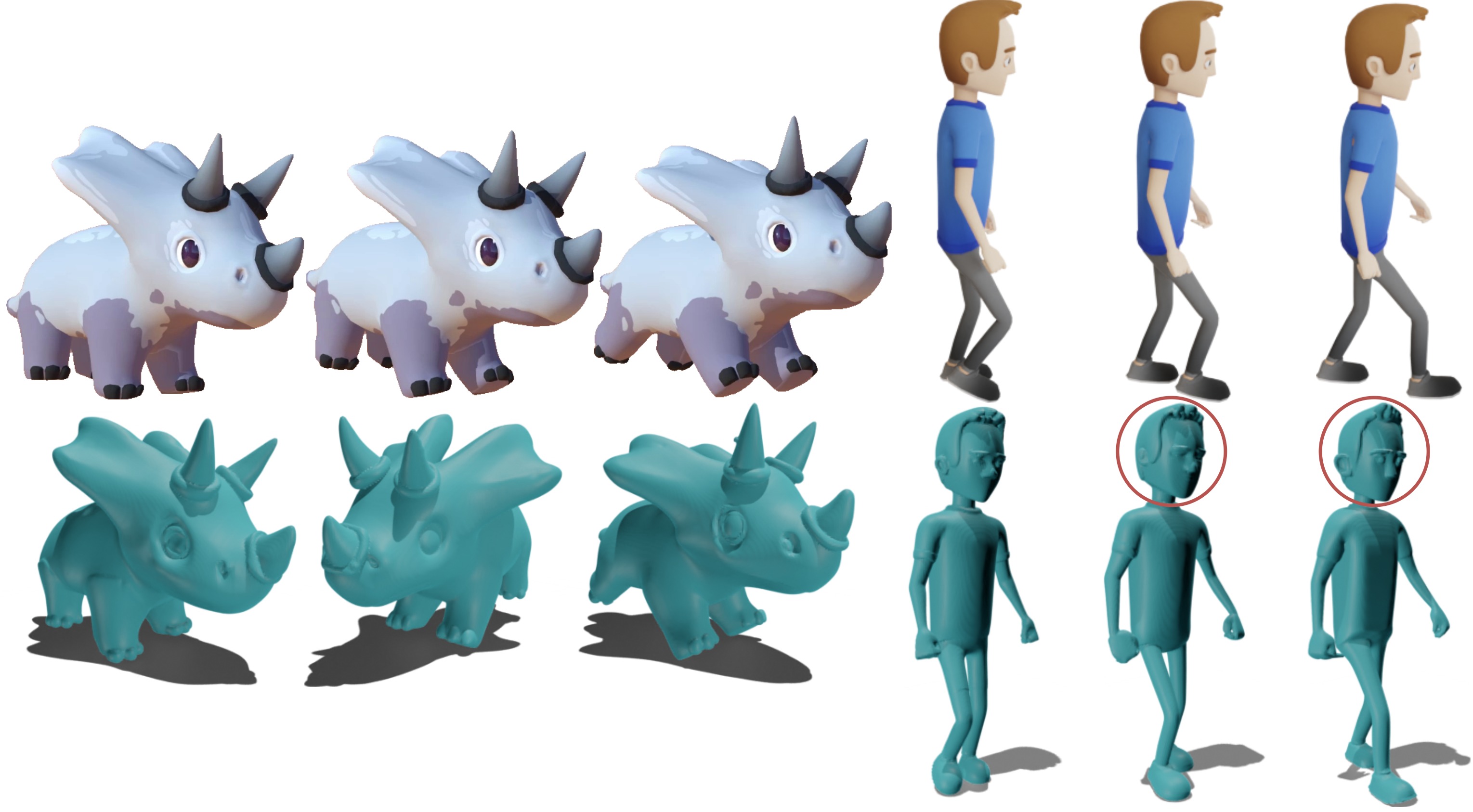}
\caption{\textbf{Image-to-3D results on video frames.} Running an image-to-3D model on each frame produces meshes exhibiting inconsistent global orientation (left) or inconsistent geometric details (right), even when using identical Gaussian noise in the denoiser.
}
\label{fig:diff_topology}
\end{figure}

Our goal in the first stage is to produce a 4D mesh, \ie, a sequence of meshes without consistent topology, from a monocular video.
A naive approach is to apply an off-the-shelf image-to-3D generator independently on all frames of the video. 
However, we found that this per-frame generation exhibits severe inconsistencies across frames, such as inconsistent 3D orientations or geometric errors that manifest as a surface flickering through time (see~\Cref{fig:diff_topology}). This is somehow expected since this naive approach lacks a mechanism enforcing consistency across frames.

To encourage cross-frame consistency, we introduce \emph{temporal 3D diffusion models}.
Drawing inspiration from multi-image models derived from pretrained image models~\citep{ho2022imagenvideo,singer2022makeavideo,shi2023mvdream,gao2024cat3d}, we propose to augment pretrained 3D diffusion models with a temporal axis to encourage the generations to be synchronized.
Specifically, we introduce two minimal changes to the original architecture, namely \emph{inflated attention} and \emph{masked generation}, which we describe next. Note that we use `temporal 3D diffusion' instead of `4D diffusion' since it is more accurate and it differentiates from recent 4D diffusion works extending image/video models to handle multi-views~\cite{watson2024fourdim,zhang2024fourdiffusion,liang2024diffusion4d}.~\Cref{fig:architecture} illustrates our resulting diffusion model architecture.

\paragraph{Inflated attention.} 

Given a sequence of $N$ frames, we build a model that outputs corresponding 3D latents in a synchronized fashion.
Inspired by MVDream~\citep{shi2023mvdream} for multi-view generation, we propose to \emph{inflate} the existing self-attention layers to allow the cross-frame synchronization of latents.
There are two benefits: (i) the tokens now attend to all tokens across frames, and (ii) we leverage existing layers that are already pretrained.
Specifically, let $\tokenfourd \in \RR^{N\times T\times D}$ be $T$ tokens of dimension $D$ corresponding to the $N$ frames.
Inflated self-attention ($\iattn$) is applied by reshaping the tensor, applying standard self-attention ($\sattn$) and reshaping back for subsequent layers as:
\begin{align}
\iattn(\tokenfourd)
=
\reshape^{-1}(\sattn(\reshape(\tokenfourd))),
\end{align}
where $\reshape$ flattens the first two dimensions such that $\reshape(\tokenfourd) \in \RR^{1\times NT\times D}$ and $\reshape^{-1}$ is the reverse operation.
To reduce the $(NT)^2$ complexity overhead, we use the efficient FlashAttention2 implementation~\citep{dao2023flashattention2}.
This mechanism improved the consistency across latents, yet, we still observed small jittering across consecutive frames.
To mitigate this issue, we propose to inject the relative frame position information via rotary positional embedding~\citep{su2023roformerenhancedtransformerrotary} inside the inflated attention layers.
We found this simple solution to generate smoother motions across frames.
Note that these simple modifications not only allow us to finetune from pretrained 3D diffusion models, but also to directly reuse the pretrained 3D autoencoders.

\paragraph{Masked generation.} The model described above generates consistent 4D output from a given video, but it does not allow us to easily start the generation from known 3D meshes, which has practical applications (see~\cref{sec:applications}). 
Hence, we turn our model into a \emph{masked generative model}~\cite{chang2022maskgit,li2022mage} where some 3D latents in the sequence are known and only the remaining `masked' latents need to be generated.
Note that such masked adaptation of pretrained generative models is reminiscent of multi-view models like CAT3D~\citep{gao2024cat3d}. 

To do so with minimal architectural changes, we propose to maintain some noise-free 3D latents during our temporal 3D diffusion model training, similar to CAT3D~\cite{gao2024cat3d}.
Specifically, let $\nsrc$ be the number of source latents and $\ntgt$ be the number of target latents such that $N = \nsrc + \ntgt$.
During training, given an input sequence of 3D latents $\latentfourd = \{\latentthreed_k\}_{k=1}^N \in \RR^{N\times T\times D}$, we randomly sample $\nsrc$ latents and keep the latter noise-free before feeding the sequence to the denoiser.
To inform the model about the noise-free latents, we set the flow matching step to 0, which is a more natural solution than CAT3D's binary mask injection.
Furthermore, we do not apply the diffusion loss on these source latents during training.
During inference, given a set of $J$ source meshes $\{\mesh_{k_j}\}_{j=1}^J$, we first use the 3D encoder $\encthreed$ to compute for each $k \in \{k_j\}_{j=1}^J$ the corresponding 3D latent $\latentthreed_{k}^\ast = \encthreed(\mesh_{k})$.
Then, before each denoising step, we copy the clean latents into the noisy latent sequence, allowing all noised tokens to attend to the latent representations of the known meshes.
Technically, our model now corresponds to a \emph{masked} temporal 3D diffusion model.

To run inference from a single video, we first use an off-the-shelf image-to-3D generator, which can be applied to any frame $\img_k$ of the video.
In particular, this allows the selection of a frame where the object is well visible and free of distortion and motion-blur.
After recovering a 3D mesh, we can then apply our masked model described above.
Note that this process is agnostic to the image-to-3D model used, therefore it would directly benefit from advances in this area. Unless stated otherwise, we use TripoSG~\citep{li2025triposg} for the image-to-3D model. See the supplemental for details. 

\subsection{Stage II: Temporal 3D Autoencoder}
\label{sec:fourd_reconstructor}

Using our temporal 3D diffusion described above, we can predict a generic 4D mesh
$\{(\verts_k,\faces_k)\}_{k=1}^{N}$ corresponding to the input video.
However, this 4D representation is impractical because the mesh topology changes throughout the sequence, preventing downstream applications such as texturing.
We thus aim at computing a 4D output corresponding to the explicit animation of a reference mesh $(\verts,\faces)$.
For this purpose, we propose to predict time-dependent vertex deformations $\deform_k$ such that $(\verts + \deform_k, \faces)$ approximates the surface of $(\verts_k, \faces_k)$.
While prior works rely on slow optimization algorithms to solve this task~\citep{sabathier2025lim, li2024dreammesh4d, chen2025v2m4}, we instead solve it with a feed-forward autoencoder taking the time-varying 3D meshes as input and outputting temporal deformation fields applied to the reference mesh.
We do so by starting from a pretrained VecSet-based VAE which we modify to handle temporal 3D data and produce deformation field outputs, hence yielding a model we term \emph{temporal 3D autoencoder}. Similar to the original VecSet VAE, such an autoencoder is also \emph{translational}, as it translates a sequence of point clouds into a sequence of deformation fields.

\paragraph{Formulation.} Given a sequence of $N$ independent meshes, we first sample 3D point clouds for each mesh and pass them independently through the frozen 3D encoder $\encthreed$ to obtain shape latents $\latentfourd = \{\latentthreed_k\}_{k=1}^N$.
Importantly, this part is identical to the original 3D autoencoder, which is critical for keeping consistency between the latents predicted by our temporal 3D diffusion and the latents of this temporal 3D autoencoder.
On the other hand, our decoder $\decfourd$ differs from the original $\decthreed$ by ingesting the entire sequence of latents $\latentfourd$ and predicting for each latent the corresponding 3D deformation field of the reference mesh.
Concretely, given two arbitrary framesteps $t_i$ and $t_j$, the decoder first processes all tokens with self-attention layers and then outputs the displacement from $t_i$ to $t_j$ of a query 3D point via a final cross-attention layer.
To indicate source and target, the framesteps $(t_i,t_j)$ are Fourier-embedded, concatenated, and injected as an extra token.
During training, the query points correspond to 3D points randomly sampled on the source mesh surface whereas during inference, we feed the 3D vertex positions of the reference mesh. In practice, we augment such query points with their normals and found that such local geometry helps disambiguate points that are spatially close yet topologically distant.
Following insights from our temporal 3D diffusion model, we encourage cross-shape consistency by inflating the decoder’s self-attention layers and encoding relative position offsets with rotary embeddings. The autoencoder is illustrated in~\Cref{fig:architecture}.

\subsection{Applications}\label{sec:applications}

\begin{figure}[t]
\centering
\includegraphics[width=0.95\linewidth]{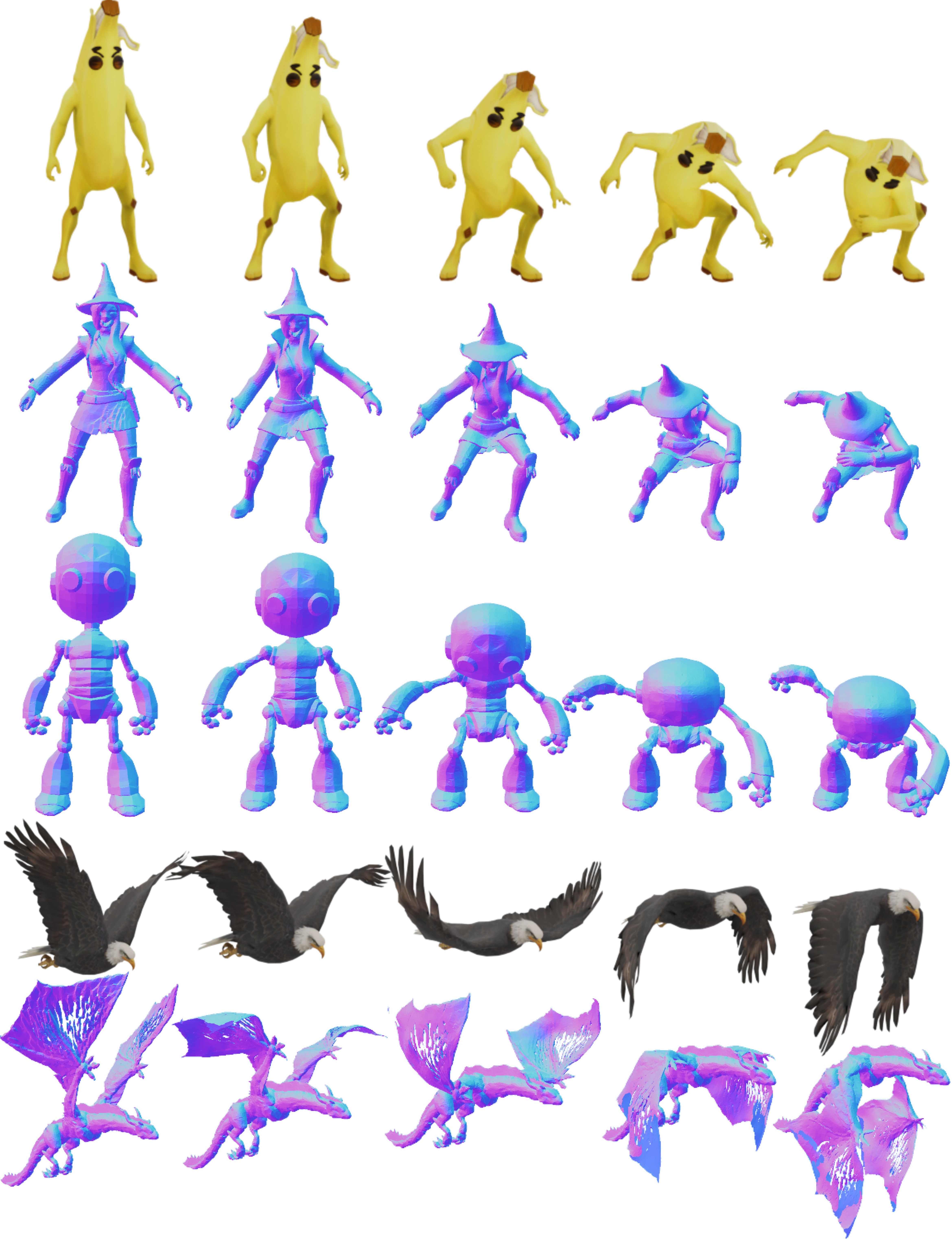}
\caption{\textbf{Motion transfer results}.
Our model is able to accurately transfer the motion from a source video to target meshes, even if the objects are inconsistent.
} 
\label{fig:motion_transfer}
\end{figure}

As described above, \name predicts an animated 3D mesh given a video, thus solving a \textbf{video-to-4D} problem.
One of its specificities is its masked generative modeling which enables us to incorporate known 3D shapes in the generation process. This characteristic not only allows us to solve a \textbf{\{3D+video\}-to-animation} problem, but also to unlock several useful applications that we describe next.~\Cref{fig:teaser,fig:motion_transfer} showcase some of these applications.

\vspace{-1em}
\paragraph{\{3D+text\}-to-animation.} To predict an animation given a mesh and a text prompt describing the motion, 
we first render the mesh using a frontal viewpoint and a white background to get an image $\img_1$.
Then, we use an off-the-shelf video model to animate the image $\img_1$ given the text description, yielding a video $\{\img_k\}_{k=1}^{N}$. Finally, we apply \name with the known 3D and the generated video.

\vspace{-1em}
\paragraph{\{Image+text\}-to-4D.} To predict an animated 3D mesh from an image depicting an object and a text prompt describing the motion, we use an off-the-shelf image-to-3D model to recover a 3D mesh, and then use our \{3D+text\}-to-animation process.

\vspace{-1em}
\paragraph{Text-to-4D.} To predict an animated 3D mesh from a single text prompt, there are two possibilities: (i) we can run a video model to generate a video from the text and use \name; or (ii) we can use an image generator to compute an image, and then run our \{image+text\}-to-animation process. In practice, we use the latter option.

\vspace{-1em}
\paragraph{Motion transfer / retargeting.} Although our model was not explicitly trained for retargeting, we found that it can transfer motion from an input video representing an object A to a different 3D object B. Specifically, this is done by simply running our \{3D+video\}-to-animation process with inconsistent objects. 

\vspace{-1em}
\paragraph{Animation extrapolation.} Given its autoregressive modeling, \name can also extrapolate animations, for instance generating coherent animations from long video sequences.
Concretely, we split the long video into chunks and use the first chunk as input to our video-to-4D component. Then, we iterate the \{3D+video\}-to-animation process on the ensuing chunks, by recursively inputting the 3D output corresponding to the last frame of the previous chunk.
\section{Experiments}
\label{sec:experiments}

We evaluate our model on the common video-to-4D task. We first compare to the state of the art through a quantitative comparison on a benchmark constructed from Objaverse~\cite{deitke_objaverse_2022} as well as a qualitative analysis on the Consistent4D benchmark~\cite{jiang_consistent4d_2023} (\Cref{sec:comp}). Then, we present additional results corresponding to other applications of our method and perform an ablation study of our key components (\Cref{sec:extra_results}).

\begin{table}[t]
\centering
\small
\caption{\textbf{Quantitative results on ActionBench.} We report results from prior works, namely LIM~\citep{sabathier2025lim}, DreamMesh4D~\citep{li2024dreammesh4d} V2M4~\citep{chen2025v2m4}, ShapeGen4D\citep{yenphraphai2025shapegen4d}, TripoSG~\citep{li2025triposg} and  TRELLIS~\citep{xiang2024structured}. We outperform all baselines across metrics while running significantly faster.}
\label{tab:main}
\addtolength{\tabcolsep}{-3pt}
\begin{tabular*}{\columnwidth}{@{\extracolsep{\fill}}lcccc@{}}
\toprule
Method & Time & $\mathrm{CD\text{-}3D}\downarrow$ & $\mathrm{CD\text{-}4D}\downarrow$ & $\mathrm{CD\text{-}M}\downarrow$ \\ 
\midrule
TRELLIS~\citep{xiang2024structured} & 15min & 0.065 & 0.181 & - \\
TripoSG~\citep{li2025triposg} & 2min & 0.056 & 0.184 & - \\
\midrule
DM4D~\citep{li2024dreammesh4d} & 35min & 0.104 &  0.152 & 0.265 \\
LIM~\citep{sabathier2025lim} & 15min & 0.089 & 0.126 & 0.243 \\
V2M4~\citep{chen2025v2m4} & 35min & 0.068 & 0.340 & 0.616 \\
SG4D~\citep{yenphraphai2025shapegen4d} & 15min & 0.056 & 0.170 & 0.348 \\
\textbf{Ours} & \textbf{2min} & \textbf{0.053} & \textbf{0.081} & \textbf{0.148} \\
\bottomrule
\end{tabular*}
\end{table}

\begin{figure*}[t]
\centering
\includegraphics[width=\linewidth]{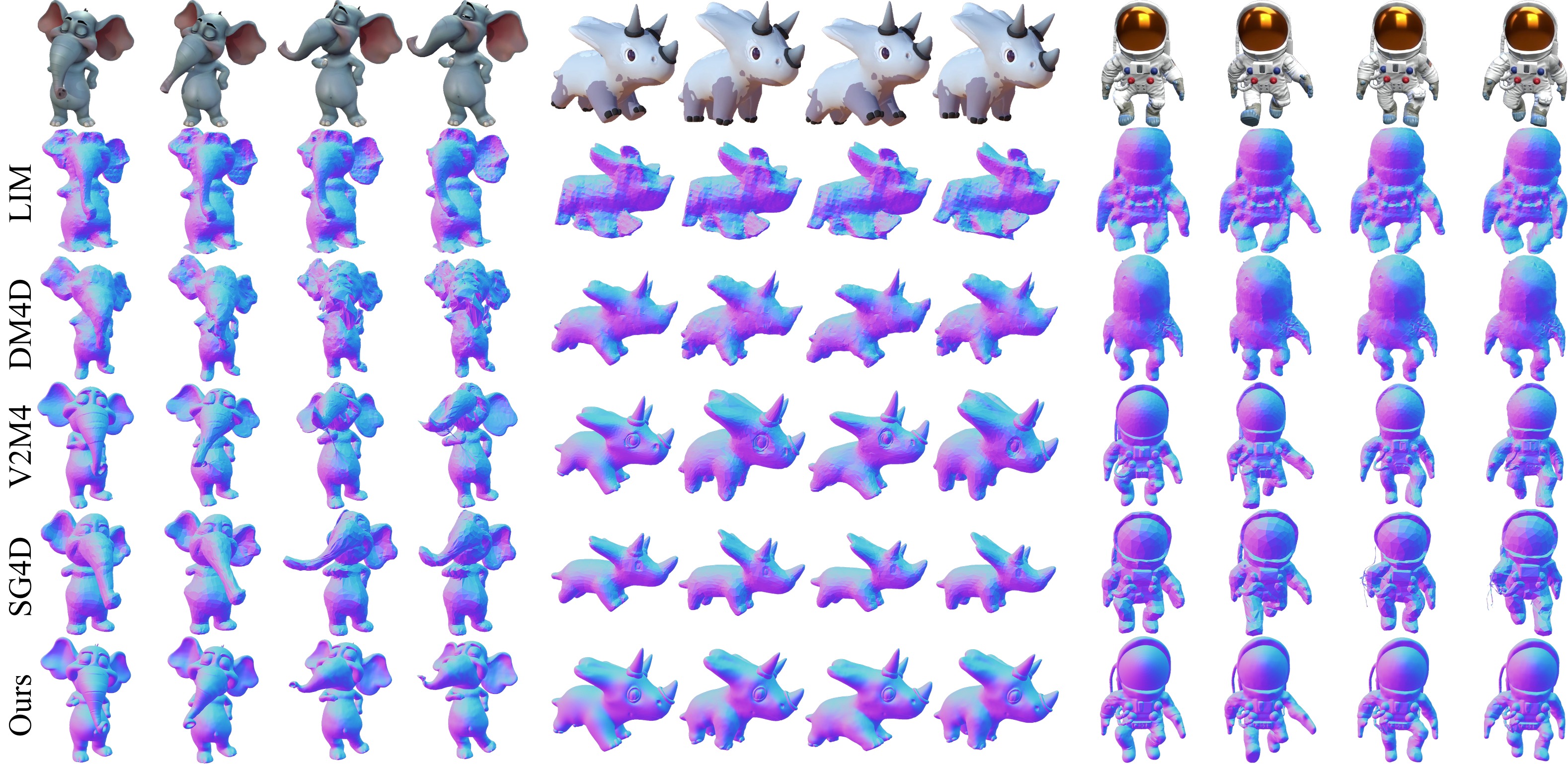}
\caption{
\textbf{Qualitative comparison on Consistent4D~\citep{jiang_consistent4d_2023}.}
LIM~\cite{sabathier2025lim} and DM4D~\cite{li2024dreammesh4d} tend to produce coarse geometries that lack details. V2M4~\cite{chen2025v2m4} and SG4D~\cite{yenphraphai2025shapegen4d} recover sharper details but leave artifacts and partial drift. In contrast, our model preserves the highest geometric fidelity across frames with a strong temporal consistency. See our project webpage for animated meshes and additional examples.
} 
\vspace*{-.1in}
\label{fig:qualitative_c4d}
\end{figure*}

\subsection{Comparison to SOTA}\label{sec:comp}

\paragraph{Baselines.} 
We compare our model to four state-of-the-art video-to-4D methods, namely LIM~\citep{sabathier2025lim}, DreamMesh4D (DM4D)~\citep{li2024dreammesh4d}, V2M4~\citep{chen2025v2m4}, and ShapeGen4D (SG4D)~\citep{yenphraphai2025shapegen4d}; and two image-to-3D models applied per-frame, TripoSG~\citep{li2025triposg} and TRELLIS~\citep{xiang2024structured}. We use the official implementations for TRELLIS, TripoSG, DM4D and V2M4. 
For LIM, we trained a re-implementation with our dataset. 
For ShapeGen4D, the authors provided results on our dataset.

\paragraph{Quantitative comparison on ActionBench.}
Since there is no open-source quantitative benchmark, we build \textit{ActionBench}, a new benchmark dataset of 128 animated scenes based on Objaverse~\citep{deitke_objaverse_2022,deitke_objaverse-xl_2023}, and release it to support reproducible evaluation and future work. We evaluate on ActionBench and report three metrics that are complementary. First, we evaluate the \emph{per-frame} 3D reconstruction quality by aligning, for each frame, the predicted mesh with ICP~\citep{BeslICP} and computing the chamfer distance between ground-truth and prediction ($\mathrm{CD\text{-}3D}$). Second, the 4D reconstruction quality is evaluated by aligning the predicted mesh sequence with a global ICP applied on the first mesh, and averaging the chamfer distance ($\mathrm{CD\text{-}4D}$). Third, we evaluate motion fidelity with a chamfer-like distance tailored to quantify motion ($\mathrm{CD\text{-}M}$). Specifically, after aligning the mesh sequence with a global ICP, we establish nearest neighbor correspondences using the first mesh. Then, for each remaining frame, we evaluate the bidirectional distance between corresponding points. We provide evaluation details in our supplementary.

We report the scores in~\Cref{tab:main}.
Quantitatively, our method outperforms the baselines across all metrics by a significant margin. Compared with the best prior result for each metric, our model improves $\mathrm{CD\text{-}3D}$, $\mathrm{CD\text{-}4D}$ and $\mathrm{CD\text{-}M}$ by 5\%, 35\%, and 39\%, respectively. In addition, it is an order of magnitude faster at inference (2min vs 15–45min for prior works, all evaluated on a single H100 GPU).

\paragraph{Qualitative comparison on Consistent4D.} 
We compare our method on videos from the standard evaluation set of Consistent4D~\cite{jiang_consistent4d_2023} and representative examples are shown in~\Cref{fig:qualitative_c4d}. We observe that LIM and DreamMesh4D exhibit reduced shape fidelity with softer details and visible artifacts. Although V2M4 and ShapeGen4D recover sharper details, they also produce artifacts and partial temporal drift. On the contrary, our model yield high-quality meshes with a better temporal coherence and a much stronger motion fidelity. It is worth noting that it does so while being significantly faster. Other examples are shown in our supplementary.

\begin{figure*}[t]
\centering
\includegraphics[width=\linewidth]{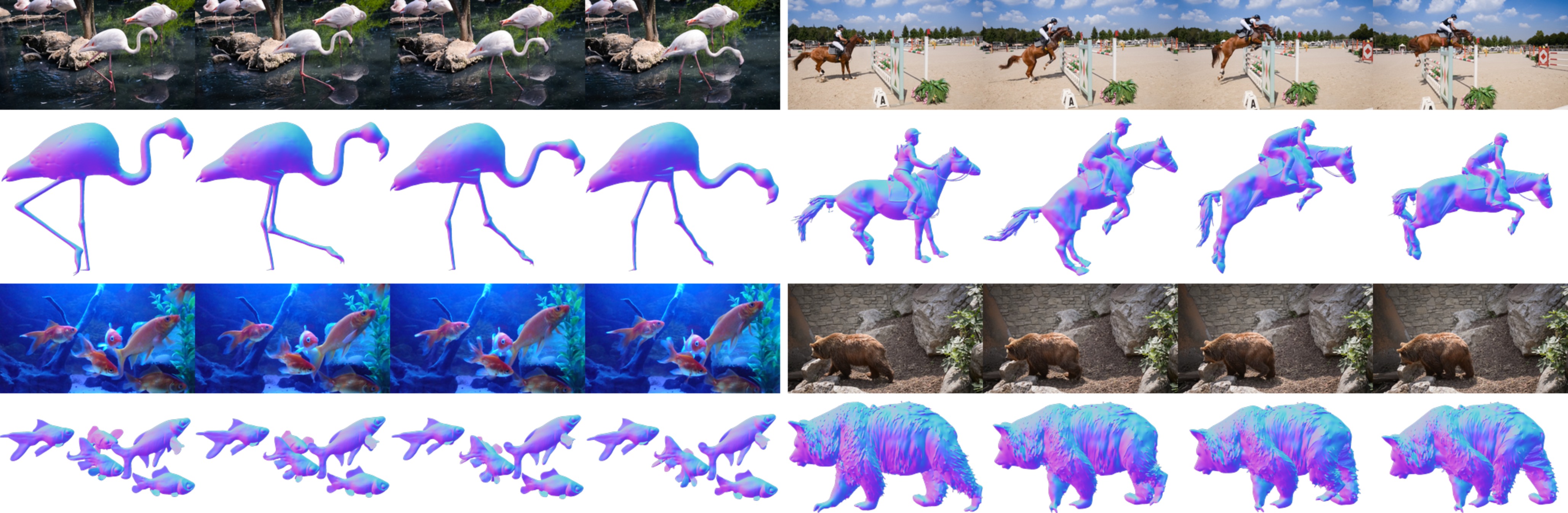}
\caption{
\textbf{Qualitative results on real videos from DAVIS~\citep{Perazzi2016DAVIS}.} Even in these challenging scenarios, our model produces accurate animated 3D meshes, thus demonstrating its ability to handle complex motions, multiple objects and occlusions. See supplemental for more results.
}
\label{fig:qualitative_davis}
\end{figure*}

\subsection{Additional results}
\label{sec:extra_results}

\paragraph{Real-world videos.}
In \Cref{fig:qualitative_davis}, we demonstrate our method's robustness by showing qualitative results on real-world videos from DAVIS~\citep{Perazzi2016DAVIS}, where a segmentation model was used to isolate the foreground object. Although our model was only trained on synthetic data, it is able to perform accurate 4D reconstructions on these challenging examples. In particular, it is able to capture large-scale motions like a jumping horse (top right) as well as more subtle movements like a bear walking (bottom right). Notably, our rig-free approach excels in handling complex, multi-part animations, as illustrated by the aquarium scene (bottom left).

\paragraph{Motion transfer.} In~\Cref{fig:motion_transfer}, we show our model's ability to transfer motion from a video representing an object A to a different object B, without explicit training for this task. We found this process to work well when semantic correspondences between objects can be established, like animating a 3D dragon using a video of a flying bird (botton).

\paragraph{Ablation study.}
\Cref{tab:ablation_main} analyzes the influence of some of our key design choices. First, we remove stage II and thus evaluate the performance of stage I only, which is not able to produce animated 3D meshes. Interestingly, stage II preserves the 3D reconstruction quality while allowing us to predict an animated mesh. Second, we remove both stages; this amounts to running the image-to-3D model (TripoSG) on each frame independently. This experiment shows that stage I is the critical component to obtain accurate 4D reconstructions. Besides, thanks to our minimal modifications, it is worth noting that adding both stages does not harm the 3D reconstruction quality of our backbone. Finally, we replace the TripoSG~\citep{li2025triposg} backbone with Craftsman~\citep{li2024craftsman} and observe that the method still achieves competitive performances. We provide additional analysis in our supplementary material.

\begin{table}[t]
\centering
\caption{\textbf{Ablation study.} We evaluate some key components on 32 animated scenes from ActionBench, namely our stage I, both stages I and II, and the pretrained model we started from by using Craftsman~\citep{li2024craftsman} instead of TripoSG~\citep{li2025triposg}.}
\label{tab:ablation_main}
\addtolength{\tabcolsep}{-2pt}
\begin{tabular}{@{}lccc@{}}
\toprule
Ablation type & $\mathrm{CD\text{-}3D}\!\downarrow$ & $\mathrm{CD\text{-}4D}\!\downarrow$ & $\mathrm{CD\text{-}M}\!\downarrow$ \\
\midrule
Full model & \textbf{0.050} & \textbf{0.069} & \textbf{0.137} \\
\quad w/o stage II & \textbf{0.050} & \textbf{0.069} & -- \\
\quad w/o stage I \& II & \textbf{0.050} & 0.187          & -- \\
\quad w/ Craftsman backbone  & 0.072 & 0.117 & 0.216 \\
\bottomrule
\end{tabular}
\vspace*{-.15in}
\end{table}
\section{Conclusion}
\label{sec:conclusion}

We presented \name, a fast, feed-forward generative model that produces animated 3D meshes that are topology-consistent and rig-free, directly from diverse inputs. Our key insight relies on temporal 3D diffusion: we extend pretrained 3D diffusion models with a temporal axis to generate a sequence of synchronized shape latents, and then use a temporal 3D autoencoder to translate these shapes into deformations of a reference mesh, yielding an animation with consistent topology. This delivers high-fidelity shape and motion in 2 minutes, enabling rapid iteration and seamless downstream use in texturing and retargeting.
We report state-of-the-art geometric accuracy and temporal consistency, demonstrating that our model is a simple, general, and practical path to production-ready animated 3D meshes.

\begin{figure}[t]
\centering
\includegraphics[width=\columnwidth]{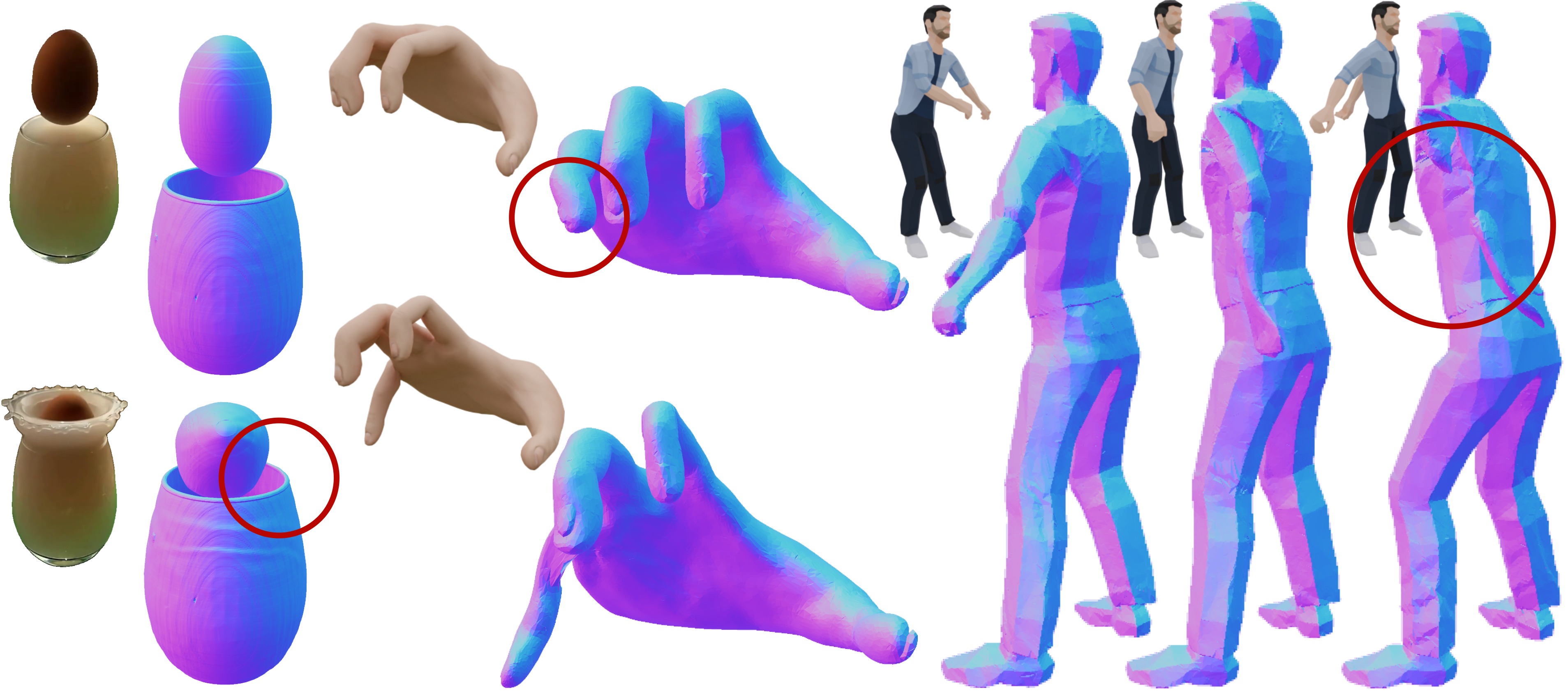}
\caption{\textbf{Limitations.}
Typical failure cases arise for videos with topological changes (left) and regions that are occluded either on the reference frame (middle) or during the motion (right).
}
\vspace*{-.15in}
\label{fig:limitations}
\end{figure}

\paragraph{Limitations and directions.} We highlight two typical failure cases in~\Cref{fig:limitations} that we discuss next:
\begin{itemize}
\item \textit{Topological changes (left).} We assume fixed connectivity and thus changes in topology cannot be modeled. Rather than explicit mesh surgery, a promising direction is to enable topology-aware latent updates that instantiate, fuse, or remove local parts without manual connectivity edits.
\item \textit{Strong occlusions (middle, right).} Although our model is able to hallucinate parts that are not visible, it sometimes fail at reconstructing occluded regions, in particular when they are missing from the reference frame or when they disappear during a complex motion.
\end{itemize}

\noindent\name's ability to lift everyday video into 4D unlocks learning geometric motion priors directly from videos. We believe this closes the loop between large-scale video corpora and mesh-native reasoning, paving the way for richer, more generalizable 4D understanding and generation.
\newpage
{
    \small
    \bibliographystyle{ieeenat_fullname}
    \bibliography{refs/main}

@String(PAMI = {IEEE Trans. Pattern Anal. Mach. Intell.})

@String(CVPR= {IEEE Conf. Comput. Vis. Pattern Recog.})

@String(ICCV= {Int. Conf. Comput. Vis.})

@String(ECCV= {Eur. Conf. Comput. Vis.})

@String(NIPS= {Adv. Neural Inform. Process. Syst.})

@String(TOG= {ACM Trans. Graph.})

@String(ICLR = {Int. Conf. Learn. Represent.})

@String(ICML = {Int. Conf. Mach. Learn.})

@String(CVPR  = {CVPR})

@String(ECCV  = {ECCV})

@String(ICCV  = {ICCV})

@String(ICLR  = {ICLR})

@String(NIPS  = {NeurIPS})

@String(PAMI  = {IEEE TPAMI})

@String(SIGGRAPH = {SIGGRAPH})

@String(SIGA = {SIGGRAPH Asia})

@String(TOG   = {ACM TOG})

@article{10.1145/37402.37422,
    title={Marching cubes: A high resolution 3D surface construction algorithm},
    author={Lorensen, William E. and Cline, Harvey E.},
    journal=SIGGRAPH,
    year={1987},
}

@article{BeslICP,
    title={A method for registration of 3-D shapes}, 
    author={Besl, P.J. and McKay, Neil D.},
    journal=PAMI, 
    year={1992},
}

@article{su2023roformerenhancedtransformerrotary,
    title={RoFormer: Enhanced Transformer with Rotary Position Embedding}, 
    author={Jianlin Su and Yu Lu and Shengfeng Pan and Ahmed Murtadha and Bo Wen and Yunfeng Liu},
    journal={Neurocomputing},
    year={2023},
}

@article{Jaegle2021PerceiverGP,
    title={Perceiver: General Perception with Iterative Attention},
    author={Andrew Jaegle and Felix Gimeno and Andrew Brock and Andrew Zisserman and Oriol Vinyals and Jo{\~a}o Carreira},
    journal=ICML,
    year={2021},
}

@article{lipman22flow,
    title={Flow Matching for Generative Modeling},
    author={Yaron Lipman and Ricky T. Q. Chen and Heli Ben-Hamu and Maximilian Nickel and Matt Le},
    journal=ICLR,
    year={2023},
}

@article{liu23flow,
    title={{Flow Straight and Fast: Learning to Generate and Transfer Data with Rectified Flow}},
    author={Xingchao Liu and Chengyue Gong and Qiang Liu},
    journal=ICLR,
    year={2023},
}

@article{oquab2023dinov2,
    title={DINOv2: Learning Robust Visual Features without Supervision},
    author={Oquab, Maxime and Darcet, Timothée and Moutakanni, Theo and Vo, Huy V. and Szafraniec, Marc and Khalidov, Vasil and Fernandez, Pierre and Haziza, Daniel and Massa, Francisco and El-Nouby, Alaaeldin and Howes, Russell and Huang, Po-Yao and Xu, Hu and Sharma, Vasu and Li, Shang-Wen and Galuba, Wojciech and Rabbat, Mike and Assran, Mido and Ballas, Nicolas and Synnaeve, Gabriel and Misra, Ishan and Jegou, Herve and Mairal, Julien and Labatut, Patrick and Joulin, Armand and Bojanowski, Piotr},
    journal={ArXiv},
    year={2023}
}

@article{li2022mage,
    title={{MAGE: MAsked Generative Encoder to Unify Representation Learning and Image Synthesis}},
    author={Li, Tianhong and Chang, Huiwen and Mishra, Shlok Kumar and Zhang, Han and Katabi, Dina and Krishnan, Dilip},
    journal=CVPR,
    year={2023}
}

@article{monnier22unicorn,
    title={{Share With Thy Neighbors: Single-View Reconstruction by Cross-Instance Consistency}},
    author={Monnier, Tom and Fisher, Matthew and Efros, Alexei A and Aubry, Mathieu},
    journal=ECCV,
    year={2022},
}

@article{siddiqui2024assetgen,
    title={{Meta 3D AssetGen: Text-to-Mesh Generation with High-Quality Geometry, Texture, and PBR Materials}},
    author={Siddiqui, Yawar and Monnier, Tom and Kokkinos, Filippos and Kariya, Mahendra and Kleiman, Yanir and Garreau, Emilien and Gafni, Oran and Neverova, Natalia and Vedaldi, Andrea and Shapovalov, Roman and Novotny, David},
    journal=NIPS,
    year={2024}
}

@article{zarzar24twinner,
    title={Twinner: {Shining} {Light} on {Digital} {Twins} in a {Few} {Snaps}},
    author={Zarzar, Jesus and Monnier, Tom and Shapovalov, Roman and Vedaldi, Andrea and Novotny, David},
    journal=CVPR,
    year={2024},
}

@article{dao2023flashattention2,
    title={Flash{A}ttention-2: Faster Attention with Better Parallelism and Work Partitioning},
    author={Dao, Tri},
    journal=ICLR,
    year={2024}
}

@article{rombach22high-resolution,
    title={High-Resolution Image Synthesis with Latent Diffusion Models},
    author={Robin Rombach and Andreas Blattmann and Dominik Lorenz and Patrick Esser and Bj{\"{o}}rn Ommer},
    journal=CVPR,
    year={2022}
}

@article{Peebles2022DiT,
    title={Scalable Diffusion Models with Transformers},
    author={William Peebles and Saining Xie},
    journal=ICCV,
    year={2023},
}

@article{chang2022maskgit,
    title={MaskGIT: Masked Generative Image Transformer},
    author={Huiwen Chang and Han Zhang and Lu Jiang and Ce Liu and William T. Freeman},
    journal=CVPR,
    year={2022}
}

@article{10.5555/3295222.3295349,
    title={Attention is all you need},
    author={Vaswani, Ashish and Shazeer, Noam and Parmar, Niki and Uszkoreit, Jakob and Jones, Llion and Gomez, Aidan N. and Kaiser, \L{}ukasz and Polosukhin, Illia},
    journal=NIPS,
    year={2017},
}

@article{deitke_objaverse_2022,
	title={Objaverse: A Universe of Annotated 3D Objects},
	author={Deitke, Matt and Schwenk, Dustin and Salvador, Jordi and Weihs, Luca and Michel, Oscar and {VanderBilt}, Eli and Schmidt, Ludwig and Ehsani, Kiana and Kembhavi, Aniruddha and Farhadi, Ali},
	journal=CVPR,
	year={2023},
}

@article{deitke_objaverse-xl_2023,
    title={Objaverse-{XL}: A Universe of 10M+ 3D Objects},
	author={Deitke, Matt and Liu, Ruoshi and Wallingford, Matthew and Ngo, Huong and Michel, Oscar and Kusupati, Aditya and Fan, Alan and Laforte, Christian and Voleti, Vikram and Gadre, Samir Yitzhak and {VanderBilt}, Eli and Kembhavi, Aniruddha and Vondrick, Carl and Gkioxari, Georgia and Ehsani, Kiana and Schmidt, Ludwig and Farhadi, Ali},
	journal=NIPS,
	year={2023},
}

@article{Perazzi2016DAVIS,
    title={A Benchmark Dataset and Evaluation Methodology for Video Object Segmentation},
    author={Perazzi, Federico and Pont-Tuset, Jordi and McWilliams, Brian and Van Gool, Luc and Gross, Markus and Sorkine-Hornung, Alexander},
    journal=CVPR,
    year={2016}
}

@article{lai2025hunyuan3d25highfidelity3d,
    title={Hunyuan3D 2.5: Towards High-Fidelity 3D Assets Generation with Ultimate Details}, 
    author={Tencent Hunyuan3D Team},
    journal={ArXiv},
    year={2025},
}

@article{li2025triposg,
    title={TripoSG: High-Fidelity 3D Shape Synthesis using Large-Scale Rectified Flow Models},
    author={Li, Yangguang and Zou, Zi-Xin and Liu, Zexiang and Wang, Dehu and Liang, Yuan and Yu, Zhipeng and Liu, Xingchao and Guo, Yuan-Chen and Liang, Ding and Ouyang, Wanli and others},
    journal=PAMI,
    year={2025},
}

@article{li2024craftsman,
    title={CraftsMan3D: High-fidelity Mesh Generation with 3D Native Generation and Interactive Geometry Refiner},
    author={Weiyu Li and Jiarui Liu and Hongyu Yan and Rui Chen and Yixun Liang and Xuelin Chen and Ping Tan and Xiaoxiao Long},
    journal=CVPR,
    year={2025},
}

@article{xiang2024structured,
    title={Structured 3D Latents for Scalable and Versatile 3D Generation},
    author={Xiang, Jianfeng and Lv, Zelong and Xu, Sicheng and Deng, Yu and Wang, Ruicheng and Zhang, Bowen and Chen, Dong and Tong, Xin and Yang, Jiaolong},
    journal=CVPR,
    year={2025},
}

@article{Chen_2025_Dora,
    title={Dora: Sampling and Benchmarking for 3D Shape Variational Auto-Encoders},
    author={Chen, Rui and Zhang, Jianfeng and Liang, Yixun and Luo, Guan and Li, Weiyu and Liu, Jiarui and Li, Xiu and Long, Xiaoxiao and Feng, Jiashi and Tan, Ping},
    journal=CVPR,
    year={2025},
}

@article{10.1145/3592442,
    title={3DShape2VecSet: A 3D Shape Representation for Neural Fields and Generative Diffusion Models},
    author={Zhang, Biao and Tang, Jiapeng and Nie\ss{}ner, Matthias and Wonka, Peter},
    journal=TOG,
    year={2023},
}

@article{zhang2024clay,
    title={CLAY: A Controllable Large-scale Generative Model for Creating High-quality 3D Assets},
    author={Longwen Zhang and Ziyu Wang and Qixuan Zhang and Qiwei Qiu and Anqi Pang and Haoran Jiang and Wei Yang and Lan Xu and Jingyi Yu},
    journal=TOG,
    year={2024},
}

@article{hong2024lrmlargereconstructionmodel,
    title={LRM: Large Reconstruction Model for Single Image to 3D}, 
    author={Yicong Hong and Kai Zhang and Jiuxiang Gu and Sai Bi and Yang Zhou and Difan Liu and Feng Liu and Kalyan Sunkavalli and Trung Bui and Hao Tan},
    journal=ICLR,
    year={2024},
}

@article{tang2024lgmlargemultiviewgaussian,
      title={LGM: Large Multi-View Gaussian Model for High-Resolution 3D Content Creation}, 
      author={Jiaxiang Tang and Zhaoxi Chen and Xiaokang Chen and Tengfei Wang and Gang Zeng and Ziwei Liu},
      journal=ECCV,
      year={2024},
}

@article{Wu2024CAT4DCA,
    title={CAT4D: Create Anything in 4D with Multi-View Video Diffusion Models},
    author={Rundi Wu and Ruiqi Gao and Ben Poole and Alex Trevithick and Changxi Zheng and Jonathan T. Barron and Aleksander Holynski},
    journal=CVPR,
    year={2024},
}

@article{Xie2024SV4DD3,
    title={SV4D: Dynamic 3D Content Generation with Multi-Frame and Multi-View Consistency},
    author={Yiming Xie and Chun-Han Yao and Vikram S. Voleti and Huaizu Jiang and Varun Jampani},
    journal={ArXiv},
    year={2024},
}

@article{Yao2025SV4D2E,
    title={SV4D 2.0: Enhancing Spatio-Temporal Consistency in Multi-View Video Diffusion for High-Quality 4D Generation},
    author={Chun-Han Yao and Yiming Xie and Vikram S. Voleti and Huaizu Jiang and Varun Jampani},
    journal={ArXiv},
    year={2025},
}

@article{xu20254dgt,
    title={4DGT: Learning a 4D Gaussian Transformer Using Real-World Monocular Videos},
    author= {Xu, Zhen and Li, Zhengqin and Dong, Zhao and Zhou, Xiaowei and Newcombe, Richard and Lv, Zhaoyang},
    journal=NIPS,
    year={2025},
}

@article{sabathier2025lim,
    title={{LIM}: Large Interpolator Model for Dynamic Reconstruction}, 
    author={Remy Sabathier and Niloy J. Mitra and David Novotny},
    journal=CVPR,
    year={2025},
}

@article{chen2025v2m4,
    title={V2M4: 4D Mesh Animation Reconstruction from a Single Monocular Video},
    author={Chen, Jianqi and Zhang, Biao and Tang, Xiangjun and Wonka, Peter},
    journal=ICCV,
    year={2025}
}

@article{li2024dreammesh4d,
    title={DreamMesh4D: Video-to-4D Generation with Sparse-Controlled Gaussian-Mesh Hybrid Representation},
    author={Zhiqi Li and Yiming Chen and Peidong Liu},
    journal=NIPS,
    year={2024}
}

@article{shi2025drive,
    title={Drive Any Mesh: 4D Latent Diffusion for Mesh Deformation from Video},
    author={Yahao Shi and Yang Liu and Yanmin Wu and Xing Liu and Chen Zhao and Jie Luo and Bin Zhou},
    journal={ArXiv},
    year={2025},
}

@article{yenphraphai2025shapegen4d,
    title={ShapeGen4D: Towards High Quality 4D Shape Generation from Videos},
    author={Jiraphon Yenphraphai and Ashkan Mirzaei and Jianqi Chen and Jiaxu Zou and Sergey Tulyakov and Raymond A. Yeh and Peter Wonka and Chaoyang Wang},
    journal={ArXiv},    
    year={2025},
}

@article{ren2024l4gm,
    title={L4GM: Large 4D Gaussian Reconstruction Model}, 
    author={Jiawei Ren and Kevin Xie and Ashkan Mirzaei and Hanxue Liang and Xiaohui Zeng and Karsten Kreis and Ziwei Liu and Antonio Torralba and Sanja Fidler and Seung Wook Kim and Huan Ling},
    journal=NIPS,
    year={2024},
}

@article{jiang_consistent4d_2023,
    title={Consistent4D: Consistent 360\{{\textbackslash}deg\} Dynamic Object Generation from Monocular Video},
	author={Jiang, Yanqin and Zhang, Li and Gao, Jin and Hu, Weimin and Yao, Yao},
	journal=ICLR,
	year={2024},
}

@article{wu_sc4d_2024,
	title={{SC}4D: Sparse-Controlled Video-to-4D Generation and Motion Transfer},
	author={Wu, Zijie and Yu, Chaohui and Jiang, Yanqin and Cao, Chenjie and Wang, Fan and Bai, Xiang},
	journal=ECCV,
	year={2024},
}

@article{yin_4dgen_2024,
	title={4DGen: Grounded 4D Content Generation with Spatial-temporal Consistency},
	author={Yin, Yuyang and Xu, Dejia and Wang, Zhangyang and Zhao, Yao and Wei, Yunchao},
	journal={ArXiv},
	year={2024},
}

@article{ren_dreamgaussian4d_2023,
	title={{DreamGaussian}4D: Generative 4D Gaussian Splatting},
	author={Ren, Jiawei and Pan, Liang and Tang, Jiaxiang and Zhang, Chi and Cao, Ang and Zeng, Gang and Liu, Ziwei},
    journal={ArXiv},
    year={2023},
}

@article{zeng_stag4d_2024,
	title={{STAG}4D: Spatial-Temporal Anchored Generative 4D Gaussians},
	author={Zeng, Yifei and Jiang, Yanqin and Zhu, Siyu and Lu, Yuanxun and Lin, Youtian and Zhu, Hao and Hu, Weiming and Cao, Xun and Yao, Yao},
	journal=ECCV,
	year={2024},
}

@article{wang2024vidu4dsinglegeneratedvideo,
    title={Vidu4D: Single Generated Video to High-Fidelity 4D Reconstruction with Dynamic Gaussian Surfels}, 
    author={Yikai Wang and Xinzhou Wang and Zilong Chen and Zhengyi Wang and Fuchun Sun and Jun Zhu},
    journal=NIPS,  
    year={2024},
}

@article{Wei2024M2V,
    title={Motion2VecSets: 4D Latent Vector Set Diffusion for Non-rigid Shape Reconstruction and Tracking},
    author={Wei Cao and Chang Luo and Biao Zhang and Matthias Nießner and Jiapeng Tang},
    journal=CVPR,
    year={2024}
}

@article{Liu2025RigAnythingTA,
    title={RigAnything: Template-Free Autoregressive Rigging for Diverse 3D Assets},
    author={Isabella Liu and Zhan Xu and Wang Yifan and Hao Tan and Zexiang Xu and Xiaolong Wang and Hao Su and Zifan Shi},
    journal=TOG,
    year={2025},
}

@article{Guo_2025_CVPR,
    title={Make-It-Animatable: An Efficient Framework for Authoring Animation-Ready 3D Characters},
    author={Guo, Zhiyang and Xiang, Jinxu and Ma, Kai and Zhou, Wengang and Li, Houqiang and Zhang, Ran},
    journal=CVPR,
    year={2025},
}

@article{Song_2025_CVPR,
    title={MagicArticulate: Make Your 3D Models Articulation-Ready},
    author={Song, Chaoyue and Zhang, Jianfeng and Li, Xiu and Yang, Fan and Chen, Yiwen and Xu, Zhongcong and Liew, Jun Hao and Guo, Xiaoyang and Liu, Fayao and Feng, Jiashi and Lin, Guosheng},
    journal=CVPR,
    year={2025},
}

@article{yao2025riggsrigging3dgaussians,
    title={RigGS: Rigging of 3D Gaussians for Modeling Articulated Objects in Videos}, 
    author={Yuxin Yao and Zhi Deng and Junhui Hou},
    journal=CVPR,
    year={2025},
}

@article{ho2022imagenvideo,
    title={Imagen Video: High Definition Video Generation with Diffusion Models},
    author={Ho, Jonathan and Chan, William and Saharia, Chitwan and Whang, Jay and Gao, Ruiqi and Gritsenko, Alexey and Kingma, Diederik P. and Poole, Ben and Norouzi, Mohammad and Fleet, David J. and Salimans, Tim},
    journal={ArXiv},
    year={2022},
}

@article{singer2022makeavideo,
    title={Make-A-Video: Text-to-Video Generation without Text-Video Data},
    author={Singer, Uriel and Polyak, Adam and Hayes, Thomas and Yin, Xi and An, Jie and Zhang, Songyang and Hu, Qiyuan and Yang, Harry and Ashual, Oron and Gafni, Oran and Parikh, Devi and Gupta, Sonal and Taigman, Yaniv},
    journal=ICLR,
    year={2023},
}

@article{shi2023mvdream,
    title={MVDream: Multi-view Diffusion for 3D Generation},
    author={Shi, Yichun and Wang, Peng and Ye, Jianglong and Mai, Long and Li, Kejie and Yang, Xiao},
    journal=ICLR,
    year={2024},
}

@article{gao2024cat3d,
    title={CAT3D: Create Anything in 3D with Multi-View Diffusion Models},
    author={Gao, Ruiqi and Holynski, Aleksander and Henzler, Philipp and Brussee, Arthur and Martin-Brualla, Ricardo and Srinivasan, Pratul and Barron, Jonathan T. and Poole, Ben},
    journal=NIPS,
    year={2024},
}

@article{watson2024fourdim,
    title={Controlling Space and Time with Diffusion Models},
    author={Watson, Daniel and Saxena, Saurabh and Li, Lala and Tagliasacchi, Andrea and Fleet, David J.},
    journal={ArXiv},
    year={2024},
}

@article{zhang2024fourdiffusion,
    title={4Diffusion: Multi-view Video Diffusion Model for 4D Generation},
    author={Zhang, Haiyu and Chen, Xinyuan and Wang, Yaohui and Liu, Xihui and Wang, Yunhong and Qiao, Yu},
    journal=NIPS,
    year={2024}, 
}

@article{liang2024diffusion4d,
    title={Diffusion4D: Fast Spatial-temporal Consistent 4D Generation via Video Diffusion Models},
    author={Liang, Hanwen and Yin, Yuyang and Xu, Dejia and Liang, Hanxue and Wang, Zhangyang and Plataniotis, Konstantinos N. and Zhao, Yao and Wei, Yunchao},
    journal=NIPS,
    year={2024}, 
}

@article{wu2025animateanymeshfeedforward4dfoundation,
      title={AnimateAnyMesh: A Feed-Forward 4D Foundation Model for Text-Driven Universal Mesh Animation}, 
      author={Zijie Wu and Chaohui Yu and Fan Wang and Xiang Bai},
      journal=ICCV,
      year={2025},
}

@article{jiang2024animate3danimating3dmodel,
      title={Animate3D: Animating Any 3D Model with Multi-view Video Diffusion}, 
      author={Yanqin Jiang and Chaohui Yu and Chenjie Cao and Fan Wang and Weiming Hu and Jin Gao},
      journal=NIPS,
      year={2024},
}

@article{muralikrishnan2025smftemplatefreerigfreeanimation,
title={SMF: Template-free and Rig-free Animation Transfer using Kinetic Codes}, 
author={Sanjeev Muralikrishnan and Niladri Shekhar Dutt and Niloy J. Mitra},
year={2025},
journal = SIGA 
}
}
\clearpage
\appendix
\setcounter{page}{1}

\twocolumn[{
\centering
\Large
\textbf{\thetitle}
\vspace{0.5em}Supplementary Material
\vspace{1.0em}
\begin{center}
\end{center}
\vspace{-0.5cm}
\vspace{0.7cm}
}]

\noindent In this supplementary document, we present additional results (\Cref{secsup:results}), more ablation studies (\Cref{secsup:ablation}) and details of implementations (\Cref{secsup:details}). We encourage readers to consult the video results provided on our webpage: \projectwebpage.


\section{Additional results}
\label{secsup:results}

\begin{figure*}[t!]
\centering
\vspace{-2em}
\includegraphics[width=0.9\linewidth]{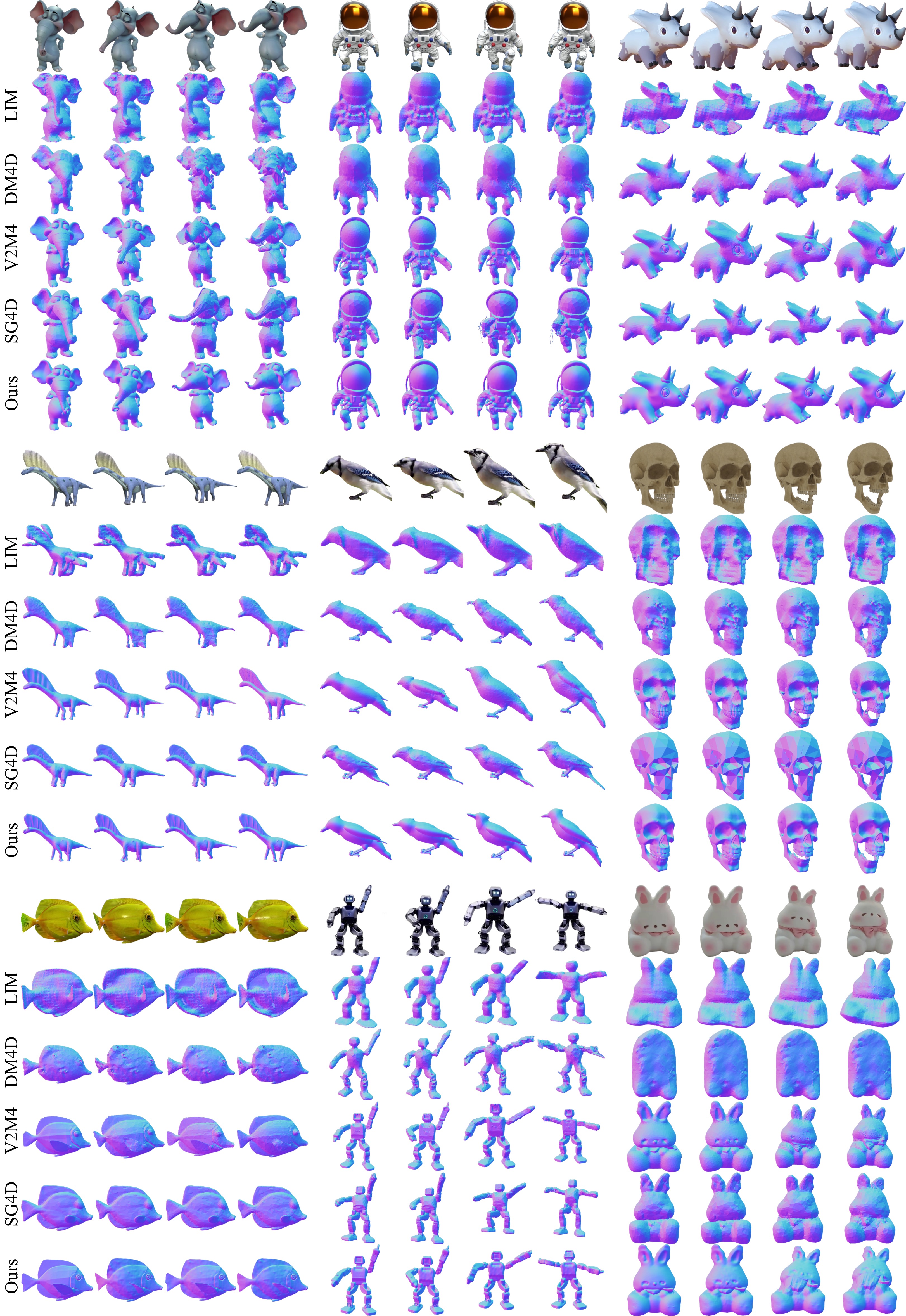}
\caption{\textbf{Additional qualitative comparison on Consistent4D~\citep{jiang_consistent4d_2023}.}}
\label{fig:qualitative_c4d_large}
\end{figure*}

\paragraph{Consistent4D comparison.}
We include additional qualitative comparisons on the standard Consistent4D~\citep{jiang_consistent4d_2023} evaluation set in \cref{fig:qualitative_c4d_large}. 
Our method produces reconstructions with sharper geometry, fewer temporal artifacts, and more faithful motion than competing approaches. 
We release videos for all Consistent4D scenes and all baselines, rendered from three viewpoints each, on the supplementary website under the section \emph{Application 1: video-to-4D}.

\begin{figure*}[t!]
\centering
\includegraphics[width=\linewidth]{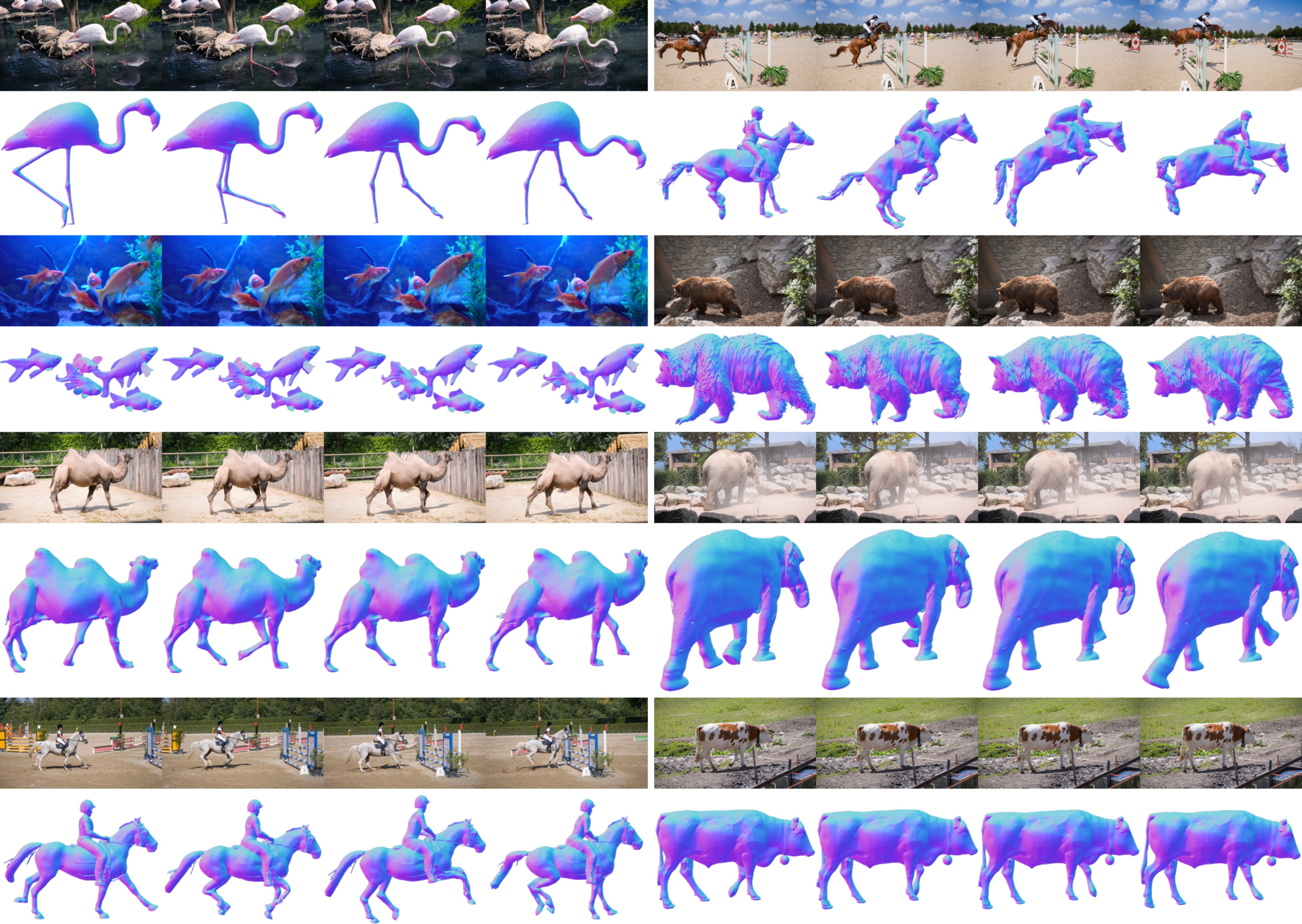}
\caption{
\textbf{Additional qualitative results on real videos from DAVIS~\citep{Perazzi2016DAVIS}.}
}
\label{fig:qualitative_davis_large}
\end{figure*}

\paragraph{DAVIS results.}
We add more qualitative results on real-world videos from DAVIS~\citep{Perazzi2016DAVIS} in \cref{fig:qualitative_davis_large}. Despite being trained exclusively on synthetic videos, our approach generalizes well to in-the-wild footage, recovering sharp geometry and plausible motion. Videos for all DAVIS examples are provided on the supplementary website under the section \emph{Application 1: video-to-4D}.

\paragraph{Long animation generation.}
In the standard setting, our model is trained to generate animated sequences of 16 frames. 
Thanks to its autoregressive modeling, we can apply our model to longer videos by recursively rolling out predictions: the last 3D output of one inference is fed back as conditioning for the next one. In the supplementary website, we showcase video-to-4D reconstructions of sequences with $61$ frames, obtained from a single standard 16-frame inference pass followed by three additional autoregressive passes. These examples illustrate that our model can maintain coherent geometry and motion over significantly extended time horizons beyond its training sequence length. Videos are listed under the section \emph{Application 1: video-to-4D}.

\paragraph{Video results.}
We provide on our supplementary website a comprehensive set of videos illustrating the five main applications of our model. 
\textit{(i) Video-to-4D:} we show reconstructions on the Consistent4D~\citep{jiang_consistent4d_2023} evaluation set and on real-world DAVIS~\citep{Perazzi2016DAVIS} videos (see paragraphs above). 
\textit{(ii) \{3D+text\}-to-4D:} given a static 3D textured mesh and a motion description, we animate well-known benchmark meshes (the armadillo and the cow) as well as additional textured meshes produced by an external 3D generative model. Since our method outputs animated meshes with fixed topology, textures remain coherent and are consistently propagated throughout the entire sequence.
\textit{(iii) \{Image+text\}-to-4D:} starting from single images drawn from standard 3D evaluation sets used in TripoSG~\citep{li2025triposg}, Dora~\citep{Chen_2025_Dora}, and Trellis~\citep{xiang2024structured}, we manually attach a short motion description to each image (\eg, the astronaut “dancing”, the mushroom “singing opera”) and generate an animated 3D mesh out of them. 
\textit{(iv) Text-to-4D:} we generate animated meshes directly from textual prompts describing both the object and its motion, such as “an octopus playing maracas”.
\textit{(v) Motion transfer:} we include motion transfer videos with two input motions, each applied on two different meshes. 

\begin{figure}[t!]
    \centering
    \includegraphics[width=\columnwidth]{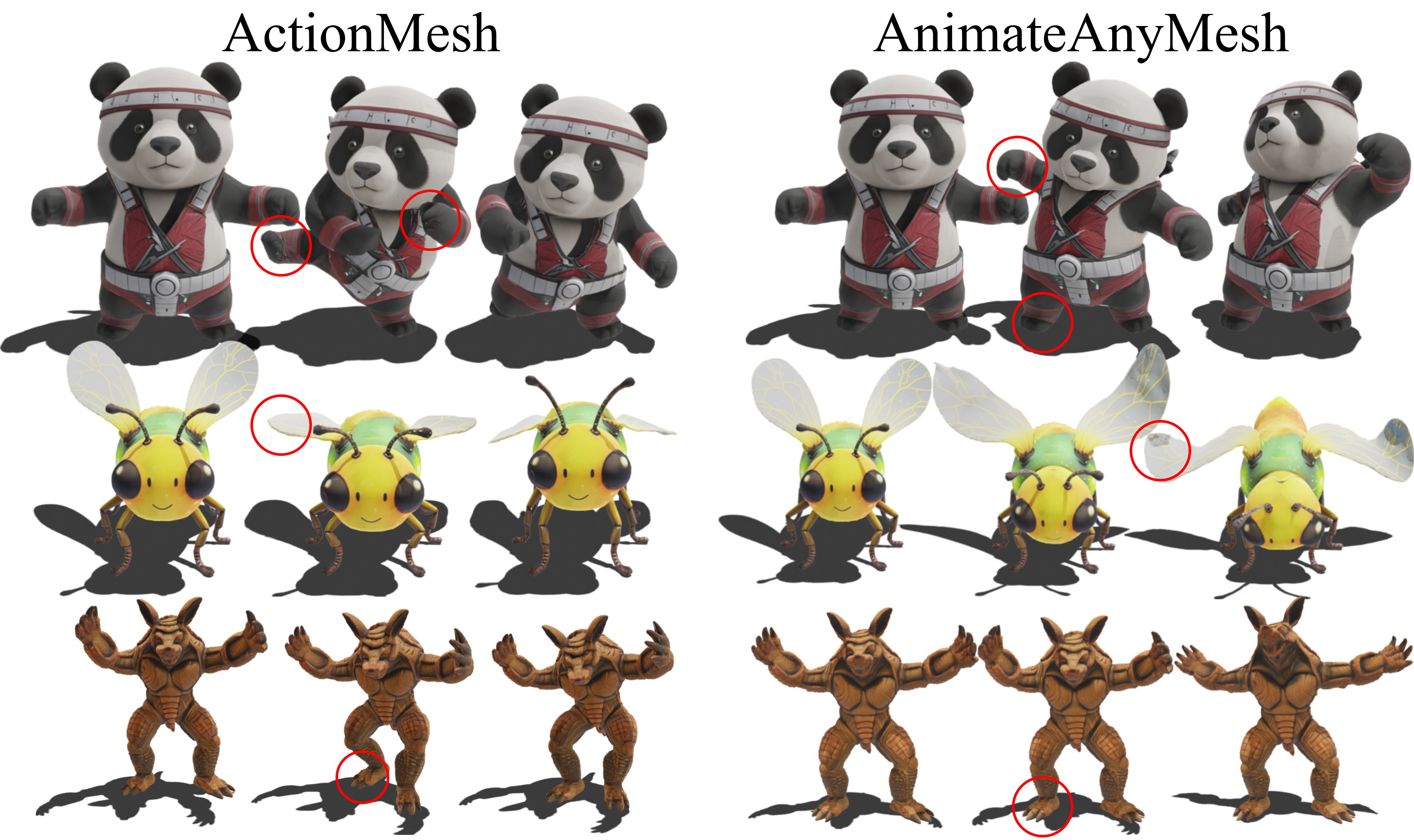}
    \caption{\textbf{Comparison to AnimateAnyMesh~\citep{wu2025animateanymeshfeedforward4dfoundation}.} Panda ``doing martial art", Firefly ``flapping its wings", Armadillo ``casually walking". The Armadillo mesh is sourced from the Stanford Computer Graphics Laboratory.}
    \label{fig:mesh_animation}
\end{figure}

\paragraph{Mesh Animation.}
As discussed in Section~\ref{sec:applications}, ActionMesh allows the integration of known 3D shapes into the generation process, enabling solutions to both the \{3D+video\}-to-animation and \{3D+text\}-to-animation problems. Several recent works have focused specifically on this task~\citep{shi2025drive,wu2025animateanymeshfeedforward4dfoundation,jiang2024animate3danimating3dmodel}.
Among these, AnimateAnyMesh~\citep{wu2025animateanymeshfeedforward4dfoundation} stands out as a text-driven, feed-forward mesh animation model capable of generating animations in approximately 6 seconds on a single NVIDIA H100 GPU.
We present a qualitative comparison between ActionMesh and AnimateAnyMesh in Figure~\ref{fig:mesh_animation}. While ActionMesh is notably slower, it produces animations with significantly greater motion expressiveness. For example, in the "panda doing martial art" scenario, ActionMesh reconstructs a coordinated motion involving both the upper and lower body, whereas AnimateAnyMesh primarily activates only the upper body. Similarly, for the "armadillo casually walking" prompt, AnimateAnyMesh fails to animate the subject's legs, while ActionMesh generates a more natural, full-body motion.
The superior expressiveness of ActionMesh stems from its reliance on the strong motion priors of video diffusion models, rather than being constrained by a limited dataset of animated objects. As described in Section~\ref{sec:applications}, our approach animates a mesh from a text prompt by first querying an off-the-shelf video generator, then applying ActionMesh to the known 3D shape and the generated video.
In contrast, AnimateAnyMesh learns its motion prior from a dataset of approximately 66k animations—primarily from animated objects in Objaverse~\citep{deitke_objaverse_2022, deitke_objaverse-xl_2023}—which inherently limits the diversity and expressiveness of the generated motions.

\begin{table}[t]
\centering
\caption{\textbf{Ablation study - Temporal 3D denoiser.}}
\label{tab:ablation_stg1}
\begin{tabular}{@{}lcc@{}}
\toprule
Ablation type & $\mathrm{CD\text{-}3D}\!\downarrow$ & $\mathrm{CD\text{-}4D}\!\downarrow$ \\
\midrule
Full model& \textbf{0.050} & \textbf{0.069} \\
\quad w/o rotary embedding & 0.054 & 0.084 \\
\quad w/o masked modeling & 0.062 & 0.116 \\
\bottomrule
\end{tabular}
\end{table}

\begin{table}[t]
\centering
\caption{\textbf{Ablation study - Temporal 3D autoencoder.}}
\label{tab:ablation_stg2}
\begin{tabular}{@{}lc@{}}
\toprule
Ablation type & $\mathrm{CD\text{-}M}\!\downarrow$ \\
\midrule
Full model& \textbf{0.137} \\
\quad w/o normals & 0.148 \\
\quad w/o $\{t_\text{src},t_\text{tgt}\}$ in self-attentions & 0.151 \\
\bottomrule
\end{tabular}
\end{table}


\section{Additional ablation studies}
\label{secsup:ablation}

\paragraph{Temporal 3D diffusion (stage I).}
We report additional ablations of the temporal 3D diffusion model in \cref{tab:ablation_stg1}. 
First, we study the importance of injecting relative frame information inside the inflated self-attention layers. 
In our default setting, we add rotary positional embedding \citep{su2023roformerenhancedtransformerrotary} of the relative frame index in the inflated self-attention layers. 
We train a variant where inflated attention is kept but all temporal rotary embeddings are removed. 
On our Objaverse evaluation set, this leads to a clear degradation of both CD-3D and CD-4D, confirming that explicit temporal encoding is critical for stable, temporally coherent 4D generation. 
Second, we ablate the masked generation mechanism. 
In the default configuration, for video-to-4D we compute a reference mesh latent $\latentthreed_1^\ast$ from an off-the-shelf image-to-3D model and inject it as a clean source latent while generating the remaining (masked) latents. 
We train a model where this mechanism is disabled and the diffusion model is conditioned only on video frames, with no noise-free latents. 
This variant cannot support several applications enabled by our masking strategy, including \{3D+text\}-to-4D, \{image+text\}-to-4D and autoregressive long-horizon generation. 
Beyond this loss of functionality, it also underperforms our full model on video-to-4D reconstruction metrics, indicating that leveraging a strong image-to-3D prior through masked modeling both broadens the application scope and improves geometric and motion fidelity.

\paragraph{Temporal 3D autoencoder (stage II).}
We report ablations of the temporal 3D autoencoder in \cref{tab:ablation_stg2}. 
First, we study the impact of augmenting query 3D points with surface normals. 
In our default configuration, each query point is represented by its position and normal, a choice motivated by the need to disambiguate deformations of points that are spatially close but topologically distant on the surface. 
Removing normals and using positions yields a drop in performance.
Second, we ablate how the source and target timesteps $(t_{\text{src}}, t_{\text{tgt}})$ that parameterize the deformation field are injected into the model. 
By default, we treat $t_{\text{src}}$ and $t_{\text{tgt}}$ as an additional token concatenated to the shape latents, so that they are jointly processed in the self-attention layers of the temporal 3D autoencoder. 
We compare this with a variant where $(t_{\text{src}}, t_{\text{tgt}})$ are instead appended as extra features to the query points (alongside position and normals). 
While this latter design has a practical advantage—enabling substantial speed-ups when predicting multiple deformation fields for the same shape tokens, since the post–self-attention context can be cached and reused—we observe a degradation in performance.

\paragraph{Autoregressive generation.}
We ablate our autoregressive design choices in \cref{tab:ablation_ar_n_frames,tab:ablation_ar_context}. 
In ~\Cref{tab:ablation_ar_n_frames}, we vary the number of keyframes $N$ used to train the temporal 3D denoiser (stage I) and the temporal 3D autoencoder (stage II). 
In the default configuration, both stages operate on sequences of 16 keyframes. 
We additionally train variants of each stage with 4 or 8 keyframes, and evaluate some pairwise combinations of (stage~I, stage~II) configurations on our evaluation set of 16-frame sequences. 
For models trained with fewer than 16 keyframes, we reconstruct the sequence by splitting it into shorter chunks and we process them autoregressively. 
We observe that having a larger sequence length is particularly important in stage~I: increasing the number of keyframes in the denoiser (autoencoder fixed) leads to a significant improvement across all metrics.
In contrast, stage~II is less sensitive to this choice.
In~\Cref{tab:ablation_ar_context}, we study the design choices of our autoregressive regime. In the standard setting, we process long videos by splitting them into consecutive chunks; the first chunk is reconstructed as usual, and for each subsequent chunk we feed the last reconstructed 3D output as additional input (reference) along with the current video segment. In practice, both the temporal 3D denoiser and the temporal 3D autoencoder can accept one or more reference meshes as input, thus defining a \emph{context window} $\cwindow$ for our model. Increasing the number of context frames, however, implicitly increases the total number of inference steps, since fewer new frames are generated per pass. On sequences of 31 timesteps, we find that enlarging the context window has limited effect on reconstruction metrics, while incurring additional computational cost. We therefore set $\cwindow=1$ in both stages in our final model to maximize efficiency.

\paragraph{Reference frame.} We ablate the impact of the reference frame in \cref{tab:ablation_reference_frame}. Specifically, we compare using the first (default), middle, or last frame as the reference for stage I and stage II. We observe that selecting either the middle or last frame as the reference leads to slightly worse quantitative results. We attribute this to the fact that the mesh is typically in a more favorable position for 3D reconstruction in the first frame, with fewer topological ambiguities and less motion blur in the reference image used by the image-to-3D model. Regardless of which reference frame is chosen, we note that our method consistently outperforms all baselines reported in the main paper.

\begin{table}[t]
\centering
\small
\caption{\textbf{Ablation study - Number of frames.} We train the temporal 3D diffusion model and the temporal 3D autoencoder with various number of frames $N$ and compare reconstructions on 16 keyframes. 
\label{tab:ablation_ar_n_frames}}
\addtolength{\tabcolsep}{-3pt}
\begin{tabular}{@{}ccccc@{}}
\toprule
$\text{Stage I -} N$ & $\text{Stage II -} N$ & $\mathrm{CD\text{-}3D}\!\downarrow$ & $\mathrm{CD\text{-}4D}\!\downarrow$ & $\mathrm{CD\text{-}M}\!\downarrow$\\
\midrule
4 & \multirow{2}{*}{16} & 0.057 & 0.091 & 0.187 \\
8 &                     & 0.055 & 0.086 & 0.176 \\
\midrule
\multirow{2}{*}{16} & 4 & 0.051 & 0.073 & 0.144 \\
                    & 8 & 0.050 & 0.073 & 0.144 \\
\midrule
4 & 4 & 0.056 & 0.091 & 0.187 \\
8 & 8  & 0.055 & 0.085 & 0.175 \\
16 & 16  & \textbf{0.050} & \textbf{0.069} & \textbf{0.137} \\
\bottomrule
\end{tabular}
\end{table}

\begin{table}[t]
\centering
\small
\caption{\textbf{Ablation study - Autoregressive context window.} We evaluate our model trained with $N = 16$ keyframes on sequences of 31 timesteps for different context window $\cwindow$.
\label{tab:ablation_ar_context}}
\addtolength{\tabcolsep}{-3pt}
\begin{tabular}{@{}ccccc@{}}
\toprule
Stage I - $\cwindow$ & Stage II - $\cwindow$ & $\mathrm{CD\text{-}3D}\!\downarrow$ & $\mathrm{CD\text{-}4D}\!\downarrow$ & $\mathrm{CD\text{-}M}\!\downarrow$ \\
\midrule
1  & \multirow{3}{*}{1} & 0.051 & 0.098 & 0.195 \\
4  &                    & 0.051 & 0.094 & 0.190 \\
8  &                    & 0.051 & 0.090 & 0.185 \\
\midrule
\multirow{3}{*}{1}  & 1 & 0.051 & 0.098 & 0.195 \\
                    & 4 & 0.050 & 0.097 & 0.196 \\
                    & 8 & 0.051 & 0.098 & 0.196 \\
\midrule
1 & 1 & 0.051 & 0.098 & 0.195 \\
4 & 4 & 0.050 & 0.094 & 0.191 \\
8 & 8 & 0.051 & 0.091 & 0.187 \\
\bottomrule
\end{tabular}
\end{table}

\begin{table}[t]
\centering
\caption{\textbf{Ablation study - Reference frame.} We evaluate our model by taking either first (default), middle or last frame as the reference frame, for stage I and stage II.}
\label{tab:ablation_reference_frame}
\begin{tabular}{@{}lccc@{}}
\toprule
Reference frame & $\mathrm{CD\text{-}3D}\!\downarrow$ & $\mathrm{CD\text{-}4D}\!\downarrow$ & $\mathrm{CD\text{-}M}\!\downarrow$ \\
\midrule
First & \textbf{0.050} & \textbf{0.069} & \textbf{0.137} \\
Middle & 0.054 & 0.074 & 0.148 \\
Last & 0.055 & 0.079 & 0.161 \\
\bottomrule
\end{tabular}
\end{table}

\section{Implementation details}
\label{secsup:details}

\paragraph{Temporal 3D diffusion.}
The temporal 3D diffusion model follows the TripoSG~\citep{li2025triposg} backbone: it is composed of 21 self-attention DiT blocks (16 heads, 2048 width, 64 latent dim).
We train with AdamW (learning rate $1\times10^{-4}$, weight decay $1\times10^{-2}$) in bfloat16 mixed precision, using a global batch size of 96 for $170{,}000$ steps.
The dataset comprises $13{,}200$ animated object sequences drawn from Objaverse~\citep{deitke_objaverse_2022}, Objaverse-XL~\citep{deitke_objaverse-xl_2023}, and an internal corpus.
\textbf{Inputs.} (i) \emph{Input frames:} for each armature-driven sequence (at least 16 and up to 128 keyframes), we render 16 viewpoints per keyframe with uniformly spaced azimuths and elevations in $[40^\circ,85^\circ]$. 
(ii) \emph{Input point-clouds:} for each sequence we construct a canonical point cloud $\pc\in\RR^{\npc\times6}$ with $\npc=500{,}000$ points (XYZ and normals) together with the armature configurations over keyframes; we then deterministically deform $\pc$ to each keyframe $k$ to obtain $\pc_k$, aligned with the rendered image $\img_k$.
Training uses 16-frame sequences: each batch contains a video clip $(16\times H\times W\times3)$ and the aligned sequence of deformed point clouds with normals $(16\times \npc\times6)$.
We extend TripoSG image pre-processing to video by computing the union object bounding box over the whole sequence and resizing every frame so that this box occupies $90\%$ of the height or width; applying a consistent crop across frames avoids spurious scale/translation cues.
Consistent with~\citep{yenphraphai2025shapegen4d}, we find that deforming a single canonical point cloud over time—rather than re-sampling points at each timestep—is critical for stable training of our temporal 3D diffusion model. 
We encode point clouds with a frozen $\encthreed$ to obtain latents; during training, we randomly select $\nsrc\in\{1,2,3\}$ latents as ground-truth conditioning and train on the remaining latents using the standard flow-matching objective.
Due to memory constraint, we train with $T=1024$ tokens, however, at inference, $T=2048$.

\paragraph{Temporal 3D autoencoder.}
The temporal 3D autoencoder model follows the TripoSG~\citep{li2025triposg} backbone. It has 16 self-attention and 1 cross-attention DiT layers (8 heads, 1024 width).
We train the temporal 3D autoencoder with the same optimization hyperparameters and bfloat16 mixed-precision setup as temporal 3D diffusion, using the identical corpus of $13{,}200$ animated object sequences. 
Training again operates on 16-frame clips: for each sequence, we provide the model with the deformed point cloud trajectory, where each point cloud contains XYZ coordinates and normals at keyframe $k$. 
In addition, we sample a source and a target timestep $(t_{\text{src}}, t_{\text{tgt}})$ uniformly between the first and last keyframe and extract a set of surface points by deforming randomly sampled mesh points from $t_{\text{src}}$ to $t_{\text{tgt}}$. 
The temporal 3D autoencoder is trained to regress these deformations from the input 3D latents on the keyframes, minimizing an $\ell_2$ loss between predicted and ground-truth positions. 
Unlike the temporal 3D diffusion stage, this training setup does not rely on latents from a temporally synchronized canonical point cloud: surface points can be re-sampled independently at each timestep, simplifying data preparation and decoupling the autoencoder from the canonical-deformation assumption.

\paragraph{Quantitative evaluation.}
Our quantitative benchmark ActionBench is composed of 128 animated scenes from Objaverse~\citep{deitke_objaverse_2022,deitke_objaverse-xl_2023}, each comprising a textured, rigged mesh with a predefined animation sequence.
For every scene we select the first $N = 16$ keyframes and render from a fixed camera with azimuth $70^\circ$.
Each mesh is converted to a watertight representation and the animation is rigidly normalized to a canonical cube $[-1,1]^3$, yielding synchronized sequences ${\{\verts_k, \faces\}}_{k=1}^{N}$ and $\{{\img_k}\}_{k=1}^{N}$.

Given watertight ground-truth meshes $\{\mesh_k\}_{k=1}^{N}$ and predicted meshes $\{\widehat{\mesh}_k\}_{k=1}^{N}$, we uniformly sample \(P = 100{,}000\) surface points from each mesh (area-weighted). 
Let $\pcs_k=\{\mathbf{x}_{k,i}\}_{i=1}^{P}\subset\RR^3$ and $\hat{\pcs}_k=\{\hat{\mathbf{x}}_{k,i}\}_{i=1}^{P}$ denote the sampled point sets. Prior to distance computation, we rigidly align predictions to the ground truth using Iterative Closest Point (ICP)~\citep{BeslICP}. 
We report two alignment protocols: \emph{ICP3D} (frame-wise), where for each $k$, we estimate $(\mathbf{r}_k, \mathbf{t}_k)\in SO(3)\times\RR^3$ that aligns $\hat{\pcs}_k$ to $\pcs_k$; and \emph{ICP4D} (sequence-wise), where we estimate $(\mathbf{r}_1,\mathbf{t}_1)$ on $k = 1$ and apply it to all frames giving $\{\mathbf{r}_1 \hat{\mathbf{x}}_{k,i}+\mathbf{t}_1\}_i$. We denote by $\bar{\pcs}_k^{\textrm{3D}}$ and $\bar{\pcs}_k^{\textrm{4D}}$  the aligned predictions under the chosen protocol.  In practice, we use the gradient-based implementation of~\cite{monnier22unicorn} to align shapes with ICP.

\smallskip\noindent
\textbf{Chamfer-based shape metrics.} 
The symmetric Chamfer distance (CD) between the ground-truth and predicted point sets at frame $k$ is defined as
\begin{equation}
\mathrm{CD}(\pcs_k,\hat{\pcs}_k)
= \frac{1}{P}\sum_{i=1}^{P}
\Big[
\min_{\hat{\mathbf{x}}\in\hat{\pcs}_k} \|\mathbf{x}_{k,i}-\hat{\mathbf{x}}\|_2^2
+
\min_{\mathbf{x}\in \pcs_k}\|
\mathbf{x} - \hat{\mathbf{x}}_{k,i}\|_2^2
\Big].
\end{equation}
We define $\mathrm{CD\text{-}3D}$ as the temporal average of the symmetric Chamfer distance computed \emph{per frame} under ICP3D alignment.
\begin{equation}
\mathrm{CD\text{-}3D}=\frac{1}{K}\sum_{k=1}^{N}\mathrm{CD}(\pcs_k,\bar{\pcs}_k^{\textrm{3D}}).
\end{equation}
When the alignment is ICP4D (single transform from \(k = 1\)), we use $\bar{\pcs}_k^{\textrm{4D}}$ in the equation above and we refer to the resulting average as $\mathrm{CD\text{-}4D}$.

\smallskip\noindent
\textbf{Motion Chamfer distance.} 
Analogous to Chamfer distance, after aligning sequence-wise using ICP4D, we establish two directed nearest-neighbor maps ($\textrm{GT}\rightarrow \textrm{PRED}$ and $\textrm{PRED}\rightarrow \textrm{GT}$ ) at the first frame and keep them fixed for the whole sequence:
\begin{align}
\sigma_i &= \argmin_{j\in\{1,\dots,P\}} \big\| \mathbf{x}_{1,i} - \hat{\mathbf{x}}_{1,j} \big\|_2^2, \\
\tau_i   &= \argmin_{j\in\{1,\dots,P\}} \big\| \mathbf{x}_{1,j}-\hat{\mathbf{x}}_{1,i} \big\|_2^2.
\end{align}

We then propagate these correspondences to every frame $k$ without recomputing nearest neighbors, and refer to the resulting average as $\textrm{CD-M}$:
\begin{align}
\textrm{CD-M}
=\frac{1}{NP}\sum_{k=1}^{N}\sum_{i=1}^{P}
\big\|\mathbf{x}_{k,i}-\hat{\mathbf{x}}_{k,\sigma_i}\big\|_2^2
+\big\|
\mathbf{x}_{k,\tau_i}-\hat{\mathbf{x}}_{k,i}\big\|_2^2.
\end{align}

\end{document}